\documentclass{article}

\PassOptionsToPackage{numbers, compress}{natbib}

 \usepackage[main, final]{neurips_2026}

\usepackage[utf8]{inputenc} 
\usepackage[T1]{fontenc}    
\usepackage{hyperref}       
\usepackage{url}            
\usepackage{booktabs}       
\usepackage{amsfonts}       
\usepackage{nicefrac}       
\usepackage{microtype}      
\usepackage{xcolor}         

\usepackage{graphicx}
\usepackage{enumitem}
\usepackage{multirow}
\usepackage{booktabs}
\usepackage[table]{xcolor}
\usepackage{graphicx}
\usepackage{amsmath}
\usepackage{wrapfig}

\usepackage[most]{tcolorbox}
\newtcolorbox{promptbox}[1]{
  colback=white,
  colframe=black,
  coltitle=white,
  colbacktitle=black,
  title=\textbf{#1},
  fonttitle=\bfseries,
  boxrule=0.8pt,
  arc=2mm,
  left=2mm,
  right=2mm,
  top=1mm,
  bottom=1mm
}

\usepackage{array}

\title{Choosing Before Acting: Comparative Value Estimation for Long-Horizon Tool-Use Agents}

\author{%
  Yu Li\textsuperscript{1},
  Zheng Zhang\textsuperscript{2},
  Xin Liu\textsuperscript{3},
  Shengtian Yang\textsuperscript{1},
  Guangfeng Cai\textsuperscript{1},
  Lei Feng\textsuperscript{1}\thanks{Corresponding author.} \\[0.5em]
  \textsuperscript{1}School of Computer Science and Engineering, 
  Southeast University, Nanjing, China \\
  \textsuperscript{2}Shandong Jianzhu University, China \\
  \textsuperscript{3}UniSA STEM, University of South Australia, Adelaide,  Australia\\
  \texttt{\{yuli11,fenglei\}@seu.edu.cn}
}

\begin{document}

\maketitle

\begin{abstract}


Large language models (LLMs) rely on long-horizon tool invocation sequences for complex tasks, where each invocation can alter the task state and condition subsequent decisions. In long-horizon tool use, final-outcome rewards provide weak credit assignment over long interaction traces. Step-level rewards can offer more targeted feedback, but obtaining reliable step supervision often requires human or LLM judgment, or additional rollouts to estimate the downstream effect of an intermediate decision.
%
In this paper, we argue that effective tool-use agents should estimate the long-horizon value of a possible next tool invocation before executing it. This objective requires comparative supervision over alternative invocations under the same context, while logged trajectories only contain the invocation that was actually taken.
Therefore, we propose \textbf{C}omparative \textbf{I}nference for \textbf{T}ool-use \textbf{A}gents (CITA). CITA trains a \textbf{C}omparative \textbf{I}nference \textbf{M}odel (CIM) from paired signals that combine observed tool behavior, scalable supervision from a Bayesian tool-graph simulator, and semantic judgments from LLM-based comparison. The resulting CIM learns to estimate how likely a possible next tool invocation is to support final task success under the current context. 
Across three tool-use benchmarks and multiple backbone LLMs, CITA consistently improves Tool F1 and task success. Additional analysis shows that CIM learns accurate step-level value estimates for comparative tool choices.

\end{abstract}

\section{Introduction}

Large language models (LLMs) extend their capabilities through external tools, including search engines \citep{nakano2021webgpt,yang2026srpo, phase}, calculators \citep{toolformer, min2, min, li2026care}, code interpreters \citep{pal}, and application programming interfaces (APIs) \citep{Gorilla}. These tools extend LLMs beyond text-only generation and enable agents to interleave reasoning with action in real tool environments \citep{react, API-Bank, ToolLLM, StableToolBench}. As tool use moves from isolated calls to long interaction traces, tool selection becomes a sequential decision problem \citep{agentbench, WebArena, AppWorld, tau}. Each tool invocation updates the task state, and an early choice can enable or block later calls \citep{toolnet, naviagent, gtool, li2026phgpo}. Therefore, the quality of a tool choice cannot be judged only by its local fit to the current instruction. It also depends on whether the choice supports final task success after the remaining interaction \citep{toolathlon, toucan,traject}.

This long-horizon dependency makes reinforcement learning (RL) difficult \citep{steptool, toolrl, toolstar}. Final-outcome rewards are easy to define, but they provide feedback only after the full trajectory is completed. For long interaction traces, such rewards give little information about which intermediate tool invocations caused success or failure \citep{salt, agentic}. 
Step-level rewards can provide more direct feedback, and process reward models have been studied for improving multi-step reasoning \citep{process, agentprm}. However, reliable step-level supervision is difficult to obtain in tool-use environments. 
Human annotation does not scale \citep{lightman2023let, steptool, toolrl}, and LLM-based judgments can be unstable \citep{agentsurvey, Judging}. Rollout-based estimation can infer the value of an intermediate decision, but it must continue the interaction from that state and execute additional tool invocations \citep{prm1, process}. Since each continuation may update the environment state, the resulting signal is costly and noisy \citep{salt}.

\begin{figure}
  \centering
  \includegraphics[width=\linewidth]{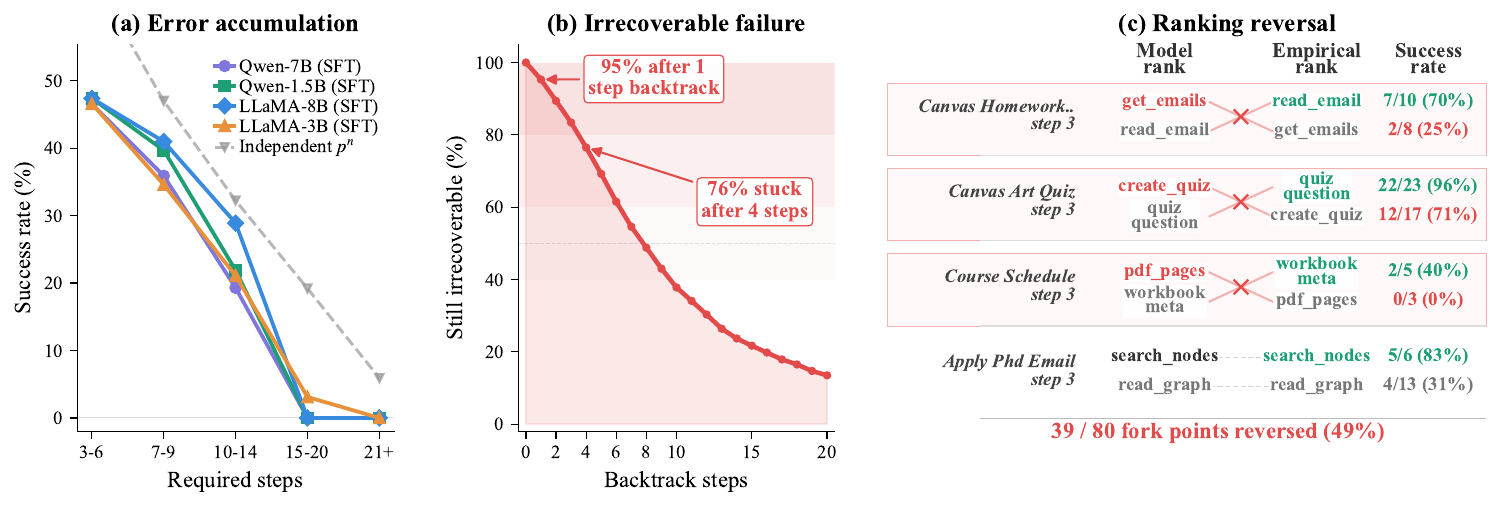}
    \caption{
    Pilot analysis on long-horizon tool-use tasks.
    (a) Task success drops sharply as the required number of tool invocations increases.
    (b) Most failed trajectories remain unrecoverable even after several backtracking steps.
    (c) At many fork points, the model-preferred tool is not the tool with the higher empirical success rate under the same context.
    These patterns motivate comparative value estimation for possible next-tool choices before execution.
    }
    \label{fig:motivation}
  \vskip -0.6cm
\end{figure}



These concerns are visible in our pilot analysis, as shown in Figure~\ref{fig:motivation}. 
The error accumulation analysis shows that task success drops sharply as the required number of tool invocations increases, indicating that errors accumulate across long interaction traces and make local tool selection unreliable. 
The failure recoverability analysis shows that 95\% of failed trajectories remain unrecoverable after one step of backtracking, and 76\% remain unrecoverable after four steps. 
This suggests that final feedback often arrives too late to guide the intermediate decisions that caused failure. 
The first two results indicate that a tool-use agent should evaluate a possible next tool invocation before execution. 
The remaining question is whether the agent's own local preference can serve as this evaluation signal. 
Our preference misalignment analysis shows that, at 49\% of fork points, the tool ranked higher by the model has a lower empirical success rate than an alternative under the same context. 
This mismatch suggests that local preference alone is not sufficient for next-tool selection.

These observations shift the objective from following local preference to estimating the downstream value of each possible next tool invocation. 
We define this signal as the long-horizon value under the current context, which measures whether a candidate invocation is likely to support final task success. 
Since this value is meaningful for decision making only relative to available choices, the agent must learn to compare alternative invocations under the same context. 
A practical obstacle is that logged trajectories only contain the invocation that was actually selected, leaving the values of unselected alternatives unobserved \citep{coma, comparative, comparative2}. 
This missing comparison signal motivates a comparative inference model that learns from paired evidence over alternative tool choices.

To instantiate this idea, we propose \textbf{C}omparative \textbf{I}nference for \textbf{T}ool-use \textbf{A}gents (CITA). 
CITA trains a \textbf{C}omparative \textbf{I}nference \textbf{M}odel (CIM) to estimate and compare the long-horizon values of candidate next-tool invocations under the current context. 
Given a tool-use context and a candidate invocation, CIM predicts whether this choice is likely to support final task success. 
To make this comparative value signal learnable, we construct paired training signals that expose alternative choices under related or shared contexts from three complementary sources: (1) Real trajectories supply pairs by matching similar decision contexts, comparing different next tools, and using their downstream outcomes as supervision. (2) A Bayesian simulator over the tool transition graph supplies scalable pairs in a sandbox, where synthetic tool success probabilities and transition reliability probabilities are sampled from Bayesian priors. This source creates controlled fork-point comparisons without executing real APIs. (3) LLM-based comparison supplies semantic pairs by comparing the factual tool choice with a plausible alternative under the same context. These complementary signals train CIM to estimate the long-horizon value of a possible next tool invocation under the current context.

CITA uses CIM as both a rollout guide and a training signal. During policy optimization training, the policy samples a set of possible next tool invocations at each decision step, and CIM selects the call with the highest predicted success value under the current context. This process produces complete trajectories guided by step-level value estimates. After execution, CIM scores the realized tool sequence by estimating \(V(s_t, a_t)\) at each step, and the aggregated scores form a comparative reward. This reward is combined with a task-level matching reward to compute group-relative advantages for policy optimization. At inference, CITA reuses the same CIM-guided selection process without policy updates. This design enables CIM to guide local tool choices during trajectory construction and provide long-horizon value feedback during policy learning.

We evaluate CITA on Toolathlon, TOUCAN, and TRAJECT-Bench with multiple backbone LLMs. CITA improves Tool F1 and end-to-end task success across benchmarks. We also evaluate CIM directly and show that it produces accurate step-level value estimates under comparative tool choices.

Our contributions are summarized as follows:
\begin{itemize}[leftmargin=0.5cm,topsep=0pt]
    \item We design a \textbf{C}omparative \textbf{I}nference \textbf{M}odel (CIM) for long-horizon tool use and train it with a three-source data construction strategy. The strategy builds paired signals from real trajectories, a Bayesian tool-graph simulator, and LLM-based comparison.

    \item We propose \textbf{C}omparative \textbf{I}nference for \textbf{T}ool-use \textbf{A}gents (CITA), a framework that uses CIM to guide policy optimization training. CIM scores sampled next-tool invocations during rollout generation and provides dense step-level comparative rewards after rollout completion.

    \item We show that CITA improves Tool F1 and end-to-end task success on three tool-use benchmarks, and that CIM learns effective step-level value estimates.
\end{itemize}

\section{Related Work}

\noindent\textbf{Prompting and Search-Based Tool Use. }
Prompting and search-based methods improve tool use by guiding LLMs to plan, call tools, and revise decisions based on observations \citep{mateval, c3tg, yang, kgerror}. 
ReAct interleaves reasoning traces with actions \citep{react}, while Toolformer trains models to decide when and how to call APIs through self-supervised data construction \citep{toolformer}. ToolLLM builds ToolBench and uses decision-tree search to collect tool-use trajectories over real APIs \citep{ToolLLM}. ToolChain* formulates API calling as action-space search and uses A* search to find efficient tool-use paths \citep{toolchain}. ART automatically composes reasoning and tool-use programs \citep{art}, ReWOO separates planning from tool execution \citep{rewoo}, Gorilla improves API calling through fine-tuning and retrieval over API documents \citep{Gorilla}, and RestGPT applies coarse-to-fine planning to real RESTful APIs \citep{song2023restgpt}.
Their main limitation lies in scalability to long-horizon settings: prompting relies on local model decisions, search becomes costly as the trajectory grows, and trace-based learning is constrained by the coverage of paths.

\noindent\textbf{Reinforcement Learning and Step-Level Rewards for Tool Use. }
Reinforcement learning methods aim to improve tool-use policies beyond supervised imitation. GRPO estimates group-relative advantages from sampled responses without training a separate critic \citep{shao2024deepseekmath}. ToolRL studies reward design for tool selection and tool application \citep{toolrl}. StepTool introduces step-grained reward shaping and optimization for multi-step tool learning \citep{steptool}. ReTool uses outcome-driven RL with real-time code execution to teach models when and how to use tools \citep{feng2025retool}. Tool-Star combines tool-integrated data synthesis with multi-tool self-critic RL for collaborative tool reasoning \citep{toolstar}. AWPO integrates explicit reasoning rewards into a GRPO-style framework through advantage weighting \citep{lin2025awpo}. PORTool builds tree-like tool-use rollouts and assigns rewards to intermediate steps for fork-relative policy optimization \citep{wu2025portool}.
These methods show that finer-grained feedback can improve tool learning. However, their step-level signals often depend on rules, annotations, LLM judgments, or rollout estimates, which can be costly or noisy in executable tool environments.

\noindent\textbf{Predictive and Comparative Planning for Interactive Agents. }
Another line of work evaluates possible actions before execution. DyMo augments function calling with state prediction in stateful tool-use environments \citep{guo2025dymo}. WebDreamer simulates candidate web actions and evaluates the simulated outcomes before acting \citep{webdreamer}. The World Knowledge Model provides global task knowledge and local dynamic knowledge for agent planning \citep{wkm}. WebCoT studies web-agent skills such as lookahead, branching, reflection, and rollback \citep{webcot}. LUMINA analyzes long-horizon interactive agents with an oracle comparative framework \citep{rakhsha2026lumina}.
These works highlight the benefit of reasoning about an action before it is executed. CITA follows this direction but focuses on the decision itself: it estimates the long-horizon value of possible next-tool invocations under the current context, without requiring full future-state prediction or oracle skills.

\section{Method}

\subsection{Problem Formulation}

We consider a long-horizon tool-use task where an LLM agent solves a user request by executing a sequence of external tool invocations. Let \(q\) denote the user query. A completed trajectory is written as \(\tau=(q,a_1,o_1,\ldots,a_T,o_T,y)\), where \(a_t\) is the tool invocation at step \(t\), \(o_t\) is the tool output, and \(y\in\{0,1\}\) indicates whether the trajectory solves the task. At step \(t\), the agent observes the current context \(s_t=(q,a_{<t},o_{<t})\), selects a possible next tool invocation \(a_t\), receives \(o_t\), and updates the context.

The goal is to choose tool invocations that maximize final task success. We define the long-horizon value of a possible next tool invocation as
\begin{equation}
V^\star(s_t,a)=\Pr(y=1\mid s_t,a),
\label{eq:long_horizon_value}
\end{equation}
which measures how likely the task is to be completed if the agent chooses \(a\) under \(s_t\). This value is not a one-step transition score. It summarizes how the current tool choice affects the remaining interaction and final outcome.

If \(V^\star(s_t,a)\) were available for all possible next tool invocations, the agent could compare alternatives under the same context and choose the one with the highest long-horizon value. In logged trajectories, however, only the executed tool invocation is observed, and the values of unselected alternatives are missing. CITA addresses this challenge by learning an estimator \(V_\phi(s_t,a)\) with a comparative inference Model (CIM). Since the target value concerns choices that are not jointly observed in a single trajectory, CIM is trained on paired signals instead of isolated trajectory labels.

\subsection{Training Signals for CIM}

Each paired signal compares two possible next-tool choices under a comparable context:
\begin{equation}
x=(s,a_f,a_c,v_f,v_c,\Delta v),
\qquad
\Delta v=v_f-v_c ,
\label{eq:paired_signal}
\end{equation}
where \(s\) is the tool-use context, \(a_f\) and \(a_c\) are two alternative next-tool invocations, and \(v_f\) and \(v_c\) are their estimated long-horizon success values. These pairs train CIM to estimate the value of each tool invocation and compare two choices under the same context.

CITA also builds a tool transition graph from logged trajectories. Each node is a tool, and each directed edge records an observed transition between two tools. The graph is used to identify plausible next-tool choices and to support the Bayesian simulator.

\noindent\textbf{\(\mathcal{D}_{\mathrm{obs}}\): observed trajectory pairs.}
The first source comes from real tool-use trajectories. CITA groups logged episodes by task and decision context, where the context includes the prefix tool sequence and compact summaries of recent arguments and outputs. For each context \(s\) and next tool \(a\), its empirical value is estimated as the downstream success rate \(\hat{v}_{\mathrm{obs}}(s,a)=N_{\mathrm{succ}}(s,a)/N(s,a)\). For two tools observed under the same context, we set \(v_f=\hat{v}_{\mathrm{obs}}(s,a_f)\), \(v_c=\hat{v}_{\mathrm{obs}}(s,a_c)\), and \(\Delta v=v_f-v_c\). This source keeps the training signal close to real tool-use behavior, but its coverage depends on the diversity of logged trajectories.

\noindent\textbf{\(\mathcal{D}_{\mathrm{sandbox}}\): Bayesian simulator pairs.}
The second source comes from a Bayesian simulator over the tool transition graph. The simulator does not execute real APIs. It creates synthetic sandbox instances where each tool node receives a sampled success probability and each transition edge receives a sampled reliability probability. For a tool node \(i\), let \(n_i\) be its call count and \(g\) be the global success rate of logged trajectories. For an edge \(i\rightarrow j\), let \(n_{ij}\) be its transition count and \(q_{ij}=n_{ij}/\sum_{j'}n_{ij'}\) be its relative frequency among outgoing transitions from \(i\). We sample
\begin{equation}
p_i\sim\mathrm{Beta}(g n_i+\lambda,(1-g)n_i+\lambda),
\qquad
r_{ij}\sim\mathrm{Beta}(q_{ij} n_{ij}+\lambda,(1-q_{ij})n_{ij}+\lambda),
\label{eq:sandbox_sampling}
\end{equation}
where \(\lambda\) is a smoothing constant. Here \(p_i\) is a synthetic success tendency used only inside the sandbox, and \(r_{ij}\) is the synthetic reliability of moving from tool \(i\) to tool \(j\).

For a continuation path \(P=(b_0,\ldots,b_L)\), its sandbox value is
\begin{equation}
\hat{v}_{\mathrm{sandbox}}(P)
=
\prod_{\ell=0}^{L}p_{b_\ell}
\prod_{\ell=0}^{L-1}r_{b_\ell b_{\ell+1}} .
\label{eq:sandbox_value}
\end{equation}
At a fork point \(u\), CITA uses the prefix ending at \(u\) as the context \(s\), selects two successor tools \(a_f\) and \(a_c\), and expands one continuation path for each successor inside the same sandbox. Let the two paths be \(P_f\) and \(P_c\). We set \(v_f=\hat{v}_{\mathrm{sandbox}}(P_f)\), \(v_c=\hat{v}_{\mathrm{sandbox}}(P_c)\), and \(\Delta v=v_f-v_c\). Since both paths are evaluated under the same sampled sandbox, the pair compares successor choices while keeping the synthetic environment fixed.

\noindent\textbf{\(\mathcal{D}_{\mathrm{cmp}}\): LLM-based comparative pairs.}
The third source uses LLM-based comparison. For each logged prefix \(s\), CITA keeps the factual next tool \(a_f\) and selects a plausible alternative \(a_c\) from valid successors or semantically related tools. The two choices are placed in the same prompt, where the LLM simulates several subsequent tool invocations and their state changes for each choice, then judges whether the task can be completed.

We repeat this paired comparison \(M\) times and keep only valid parsed outputs. Let \(M_s\) be the number of valid comparisons, and let \(y_f^{(m)},y_c^{(m)}\in\{0,1\}\) denote whether the LLM judges the simulated continuations after \(a_f\) and \(a_c\) as successful in the \(m\)-th valid comparison. We estimate \(\hat{v}_{\mathrm{cmp}}(s,a_f)=M_s^{-1}\sum_{m=1}^{M_s}y_f^{(m)}\) and \(\hat{v}_{\mathrm{cmp}}(s,a_c)=M_s^{-1}\sum_{m=1}^{M_s}y_c^{(m)}\).
Pairs are discarded when too few comparisons are parsed or when repeated judgments show low consistency. This source provides semantic comparison signals for plausible alternatives that are rare in logged trajectories or hard to cover with the simulator alone.

The three sources serve complementary roles. \(\mathcal{D}_{\mathrm{obs}}\) anchors CIM to real tool behavior, \(\mathcal{D}_{\mathrm{sandbox}}\) provides scalable structured comparisons, and \(\mathcal{D}_{\mathrm{cmp}}\) adds semantic comparison signals. CITA uses phase-aware mixing: early training relies more on simulator pairs, middle training balances all sources, and later training increases the weight of observed and LLM-based pairs to better match real tool-use contexts.

\begin{figure}[t]
  \centering
  \includegraphics[width=\linewidth]{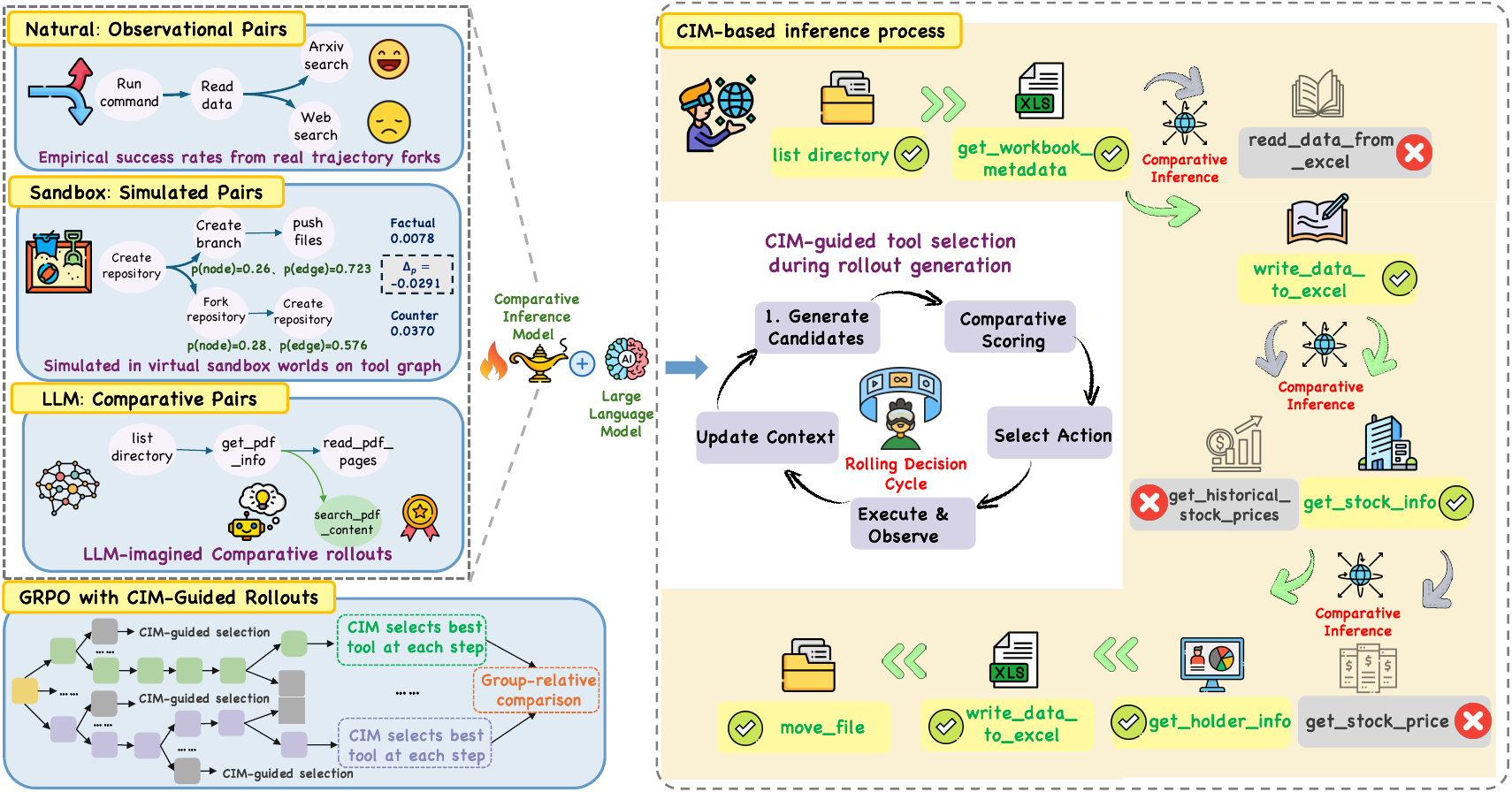}
  \caption{
  Overview of CITA. CITA trains a Comparative Inference Model (CIM) from paired signals built from real trajectories, a Bayesian tool-graph simulator, and LLM-based comparison. The trained CIM guides GRPO rollouts, provides dense comparative rewards, and supports iterative next-tool selection during inference.
  }
  \label{fig:method}
\end{figure}

\subsection{Architecture of the Comparative Inference Model}

The role of CIM is to estimate the long-horizon value of a possible next tool invocation under the current tool-use context. Given a context \(s\) and a candidate tool invocation \(a\), it outputs a value estimate \(V_\phi(s,a)\in[0,1]\) and a confidence score \(c_\phi(s,a)\in[0,1]\). The value estimate predicts how likely choosing \(a\) under \(s\) is to support final task success, while the confidence score measures the reliability of this estimate.

The model first encodes the textual information in the tool-use state, including the user query, previous tool invocations, tool outputs, and the candidate tool. We use the LLM backbone as a semantic encoder and project its hidden states into a lower-dimensional space for value estimation. A context encoder summarizes the current tool-use state into \(h_s\), and an action encoder maps the candidate tool into \(h_a\).

A scoring module combines \(h_s\) and \(h_a\) into a decision representation:
\begin{equation}
z_{s,a}=g_\phi(h_s,h_a),
\qquad
V_\phi(s,a)=\sigma(f_v(z_{s,a})),
\qquad
c_\phi(s,a)=\sigma(f_c(z_{s,a})) .
\label{eq:cim_output}
\end{equation}
Here \(g_\phi\) is the scoring module, \(f_v\) and \(f_c\) are the value and confidence heads, and \(\sigma(\cdot)\) is the sigmoid function. The two heads use the same decision representation but learn different targets. In implementation, the confidence head reads a detached representation, so confidence calibration does not directly interfere with value learning.

CIM uses paired forward computation during training. For a paired signal \(x=(s,a_f,a_c,v_f,v_c,\Delta v)\), the context \(s\) is encoded once and shared by both choices:
\begin{equation}
\hat{v}_f=V_\phi(s,a_f),
\qquad
\hat{v}_c=V_\phi(s,a_c),
\qquad
\Delta\hat{v}=\hat{v}_f-\hat{v}_c .
\label{eq:paired_forward}
\end{equation}
This design lets the model learn both the value of each possible next tool invocation and the value gap between two choices under the same context. At inference time, CIM uses the single-call form \(V_\phi(s,a)\) and \(c_\phi(s,a)\) to score candidate tool invocations.

\subsection{CIM Training}

CIM is trained to learn two abilities: estimating the value of a possible next tool invocation and comparing two choices under the same context. For a pair \(x=(s,a_f,a_c,v_f,v_c,\Delta v)\), CIM produces \(\hat{v}_f\), \(\hat{v}_c\), and \(\Delta\hat{v}\) as defined in Eq.~\ref{eq:paired_forward}. We first supervise the value estimates with
\begin{equation}
\mathcal{L}_{\mathrm{value}}
=
\frac{1}{2}
\left[
\mathrm{BCE}(\hat{v}_f,v_f)
+
\mathrm{BCE}(\hat{v}_c,v_c)
\right].
\label{eq:value_loss}
\end{equation}
This term teaches CIM to predict the long-horizon success value of each tool choice.

To make pairwise comparison explicit, we further match the predicted value gap to the paired label:
\begin{equation}
\mathcal{L}_{\mathrm{gap}}
=
(\Delta\hat{v}-\Delta v)^2 .
\label{eq:gap_loss}
\end{equation}
This term is central to comparative value estimation, because the agent must decide which choice is better under the same context.

We add two auxiliary objectives. The consistency loss \(\mathcal{L}_{\mathrm{cons}}\) regularizes value gaps with self-identity and transitivity constraints. The confidence loss \(\mathcal{L}_{\mathrm{conf}}\) calibrates the confidence head using pseudo-labels derived from prediction errors, so that low-error estimates receive higher confidence.

The final objective is
\begin{equation}
\mathcal{L}_{\mathrm{CIM}}
=
\mathcal{L}_{\mathrm{value}}
+
\alpha \mathcal{L}_{\mathrm{gap}}
+
\beta \mathcal{L}_{\mathrm{cons}}
+
\gamma \mathcal{L}_{\mathrm{conf}} .
\label{eq:cim_loss}
\end{equation}
We use a progressive training schedule. CIM first learns stable value prediction, then incorporates pairwise gap and consistency objectives, and finally calibrates confidence for rollout guidance and reward aggregation.

\subsection{CIM-Guided GRPO and Inference}

CITA uses CIM as both a rollout guide and a reward model for policy optimization. During GRPO training, the policy samples a set of possible next-tool invocations \(\mathcal{A}_t\) at each decision step. CIM scores each candidate under the current context by combining its predicted value and confidence:
\begin{equation}
S_\phi(s_t,a)=V_\phi(s_t,a)c_\phi(s_t,a).
\label{eq:cim_score}
\end{equation}
The tool invocation with the highest score is selected and executed:
\begin{equation}
a_t=\arg\max_{a\in\mathcal{A}_t}S_\phi(s_t,a).
\end{equation}
The resulting output is appended to the context, and the process continues until the rollout terminates. This procedure lets CIM guide local tool choices while the policy still generates the candidate set.


For policy optimization, CITA aggregates the CIM scores of the executed tool sequence into a comparative reward. For each executed step \(t\), CIM evaluates the realized context \(s_t\) and the selected tool invocation \(a_t\), producing \(V_\phi(s_t,a_t)\) and \(c_\phi(s_t,a_t)\). These step-level estimates can be obtained during rollout generation or by a lightweight scoring pass over the completed trajectory.
\begin{equation}
r_{\mathrm{cmp}}(\tau)
=
\frac{\sum_{t=1}^{T}c_\phi(s_t,a_t)V_\phi(s_t,a_t)}
{\sum_{t=1}^{T}c_\phi(s_t,a_t)+\epsilon}.
\label{eq:cmp_reward}
\end{equation}
This reward gives larger weight to steps where CIM is more confident. The final reward combines this comparative reward with a task-level matching reward:
\begin{equation}
r(\tau)
=
w_{\mathrm{task}}r_{\mathrm{task}}(\tau)
+
w_{\mathrm{cmp}}r_{\mathrm{cmp}}(\tau).
\label{eq:total_reward}
\end{equation}
The combined reward is used to compute group-relative advantages in GRPO. In this way, CITA provides a step-aware training signal while still optimizing complete tool-use trajectories.

At inference time, CITA applies the same CIM-guided selection rule without updating the policy. At each step, the policy samples possible next-tool invocations, and CIM scores each candidate using Eq.~\ref{eq:cim_score}. The agent executes the highest-scoring invocation, appends the returned output to the context, and repeats this process until the task is completed or a stopping condition is met.

\section{Experiments}





\subsection{Setup}

\noindent\textbf{Benchmarks.}
We evaluate CITA on three long-horizon tool-use benchmarks: Toolathlon \citep{toolathlon}, TOUCAN \citep{toucan}, and TRAJECT-Bench \citep{traject}. 
These benchmarks cover different types of tool-use difficulty. 
Toolathlon focuses on long-horizon dependency with about 13.5 tool invocations per episode. 
TOUCAN has shorter trajectories, about 3.2 calls per episode, but a large tool pool of 15,294 unique tools. 
TRAJECT-Bench provides a medium-horizon setting with about 6.4 tool invocations per episode and 715 unique tools. 
Appendix~\ref{app:dataset_stats} provides detailed statistics.

\noindent\textbf{Backbone models.}
We report results with Qwen2.5-7B \citep{qwen2} and Llama3.1-8B \citep{llama}. 
Both models are first adapted to the tool-use format with supervised fine-tuning, and then optimized with the corresponding training method. 

\noindent\textbf{Baselines.}
We compare CITA with baselines from three categories. 
The first category includes prompting and search-based tool-use methods, including ReAct \citep{react}, ToolChain* \citep{toolchain}, and ToolLLM \citep{ToolLLM}. 
The second category includes reinforcement learning and step-level reward methods, including ToolRL \citep{toolrl}, StepTool \citep{steptool}, AWPO \citep{lin2025awpo} , IPR \citep{ipr}, and AgentPRM \citep{agentprm}. 
The third category includes predictive and planning-based methods, including DyMo \citep{guo2025dymo}, WebDreamer \citep{webdreamer}, and WKM \citep{wkm}. 

\noindent\textbf{Metrics.}
We use \textbf{Tool F1} and \textbf{Task Success Rate} as the primary metrics. 
Tool F1 evaluates whether the generated trajectory selects the correct tools compared with reference trajectories, while Task Success Rate evaluates whether the completed trajectory solves the task. 
The former reflects tool-selection quality, and the latter reflects end-to-end task completion.
For Task Success Rate, we send the full tool-use process, including tool invocations and observations, to GPT-4o and use the same evaluation prompt for all methods.

\subsection{Main Results}

\begin{table*}[t]
\centering
\setlength{\tabcolsep}{3.2pt}
\renewcommand{\arraystretch}{1.08}
\caption{Main results on three benchmarks. We report Tool F1 (\%) and Task Success Rate (TSR, \%) with Qwen2.5-7B and Llama3.1-8B as backbone models. The best results are highlighted in bold.}
\label{tab:main_results}
\resizebox{\textwidth}{!}{
\begin{tabular}{lcccccccccccc}
\toprule
\multirow{2}{*}{Method}
& \multicolumn{4}{c}{Toolathlon}
& \multicolumn{4}{c}{TOUCAN}
& \multicolumn{4}{c}{TRAJECT-Bench} \\
\cmidrule(lr){2-5}\cmidrule(lr){6-9}\cmidrule(lr){10-13}
& \multicolumn{2}{c}{Qwen2.5-7B}
& \multicolumn{2}{c}{Llama3.1-8B}
& \multicolumn{2}{c}{Qwen2.5-7B}
& \multicolumn{2}{c}{Llama3.1-8B}
& \multicolumn{2}{c}{Qwen2.5-7B}
& \multicolumn{2}{c}{Llama3.1-8B} \\
\cmidrule(lr){2-3}\cmidrule(lr){4-5}
\cmidrule(lr){6-7}\cmidrule(lr){8-9}
\cmidrule(lr){10-11}\cmidrule(lr){12-13}
& Tool F1\(\uparrow\) & TSR\(\uparrow\)
& Tool F1\(\uparrow\) & TSR\(\uparrow\)
& Tool F1\(\uparrow\) & TSR\(\uparrow\)
& Tool F1\(\uparrow\) & TSR\(\uparrow\)
& Tool F1\(\uparrow\) & TSR\(\uparrow\)
& Tool F1\(\uparrow\) & TSR\(\uparrow\) \\
\midrule

\rowcolor{gray!15}
\multicolumn{13}{l}{\textit{Prompting and search-based tool use}} \\
ReAct 
& 33.00 & 22.73 & 43.79 & 25.41
& 42.74 & 32.68 & 44.00 & 34.22
& 43.64 & 30.96 & 33.71 & 23.58 \\
ToolChain*
& 38.21 & 27.84 & 42.22 & 26.17
& 25.59 & 16.93 & 48.11 & 37.26
& 41.97 & 29.74 & 47.85 & 36.77 \\
ToolLLM
& 21.10 & 11.62 & 35.43 & 20.43
& 26.28 & 17.59 & 33.34 & 27.11
& 50.10 & 31.50 & 54.68 & 32.00 \\

\midrule
\rowcolor{gray!15}
\multicolumn{13}{l}{\textit{Reinforcement learning and step-level rewards}} \\
ToolRL
& 28.46 & 16.35 & 40.95 & 26.49
& 30.75 & 21.00 & 32.00 & 22.86
& 45.46 & 32.29 & 56.38 & 40.62 \\
StepTool
& 38.64 & 24.57 & 45.32 & 30.50
& 33.79 & 20.84 & 37.03 & 22.00
& 40.11 & 29.55 & 45.75 & 32.10 \\
IPR
& 35.55 & 21.75 & 28.86 & 20.64
& 37.81 & 25.00 & 19.40 & 17.65
& 55.02 & 43.18 & 43.52 & 38.20 \\
AWPO
& 34.82 & 20.90 & 42.46 & 27.54
& 40.06 & 30.25 & 38.46 & 24.25
& 43.16 & 32.69 & 35.89 & 25.18 \\
AgentPRM
& 32.11 & 24.67 & 28.42 & 24.50
& 34.84 & 28.30 & 10.63 & 12.00
& 39.30 & 27.75 & 25.10 & 18.34 \\

\midrule
\rowcolor{gray!15}
\multicolumn{13}{l}{\textit{Predictive and planning-based methods}} \\
DyMo
& 44.55 & 28.45 & 37.13 & 22.50
& 42.26 & 28.55 & 38.11 & 26.36
& 46.63 & 30.08 & 32.66 & 22.31 \\
WebDreamer
& 47.97 & 32.74 & 55.38 & 38.41
& 27.68 & 26.76 & 29.73 & 27.35
& 48.01 & 37.25 & 61.69 & 48.16 \\
WKM
& 39.39 & 26.13 & 47.32 & 35.50
& 32.99 & 23.15 & 49.95 & 38.62
& 53.38 & 42.05 & 66.47 & 53.00 \\

\midrule
\rowcolor{blue!8}
\textbf{CITA (Ours)}
& \textbf{58.56} & \textbf{47.82}
& \textbf{59.29} & \textbf{45.05}
& \textbf{59.67} & \textbf{54.23}
& \textbf{56.44} & \textbf{47.08}
& \textbf{59.30} & \textbf{48.31}
& \textbf{69.58} & \textbf{55.72} \\

\bottomrule
\end{tabular}
}
\end{table*}


\noindent\textbf{Comparison with prior methods.}
Table~\ref{tab:main_results} shows that CITA achieves the best results in all six dataset-backbone settings on both Tool F1 and TSR, with average gains of 7.55 and 9.93 points over the strongest prior method. The improvement in Tool F1 shows that CITA generates tool-use trajectories that better match high-quality reference paths. More importantly, the larger gain in TSR shows that CITA does not simply imitate reference trajectories in a rigid way. It also improves the chance that the generated trajectory completes the task. This is important for tool-use agents, because multiple valid tool paths may solve the same task.

The advantage becomes clearer when comparing different groups of baselines. Prompting and search-based methods can find useful paths in selected settings, such as ReAct on TOUCAN and ToolLLM on TRAJECT-Bench, but their performance varies across datasets and backbones. RL and step-level reward methods improve some cases, with IPR reaching 55.02 Tool F1 and 43.18 TSR on TRAJECT-Bench with Qwen2.5-7B, yet they still fall behind CITA on both metrics. Predictive and planning-based methods are the strongest prior group, especially WebDreamer on Toolathlon and WKM on TRAJECT-Bench. CITA still improves over them, which suggests that directly estimating the long-horizon value of next-tool choices gives a more reliable decision signal than local prompting, generic step rewards, or full future-state prediction.

\noindent\textbf{Performance across benchmark settings.}
CITA performs consistently across different benchmark profiles. 
It improves TSR on Toolathlon, where long trajectories amplify early mistakes; gives the largest gains on TOUCAN, where the large tool pool makes selection difficult; and remains best on TRAJECT-Bench, where planning-based baselines are already strong. 
These results show that comparative value estimation helps under long-horizon dependency, large action spaces, and trajectory-level evaluation. 
Training dynamics are provided in Appendix~\ref{app:reward_curves}.

\subsection{CIM Value Estimation Quality}

We next evaluate whether CIM learns effective comparative value estimates for tool choices. 
Given a held-out paired example, CIM predicts the value gap \(\Delta\hat{v}\) between the factual tool choice and an alternative tool choice. 
We evaluate this prediction from three aspects. 
Value Gap MAE measures the absolute error between \(\Delta\hat{v}\) and the observed value gap \(\Delta v\). 
Direction Accuracy measures whether the sign of \(\Delta\hat{v}\) matches the sign of \(\Delta v\), which indicates whether CIM selects the better tool choice. 
AUROC measures whether \(|\Delta\hat{v}|\) separates strong-effect pairs from near-neutral pairs.

\begin{table}[t]
\centering
\small
\setlength{\tabcolsep}{3.2pt}
\renewcommand{\arraystretch}{1.05}
\caption{CIM value estimation quality on held-out paired examples.}
\label{tab:cim_quality}
\resizebox{\linewidth}{!}{
\begin{tabular}{lccccccccc}
\toprule
\multirow{2}{*}{Model}
& \multicolumn{3}{c}{TOUCAN}
& \multicolumn{3}{c}{Toolathlon}
& \multicolumn{3}{c}{TRAJECT-Bench} \\
\cmidrule(lr){2-4}\cmidrule(lr){5-7}\cmidrule(lr){8-10}
& MAE \(\downarrow\) & Dir. \(\uparrow\) & AUROC \(\uparrow\)
& MAE \(\downarrow\) & Dir. \(\uparrow\) & AUROC \(\uparrow\)
& MAE \(\downarrow\) & Dir. \(\uparrow\) & AUROC \(\uparrow\) \\
\midrule
CIM w/ \(\mathcal{L}_{\text{value}}\) only
& 0.4752 & 0.6923 & 0.9615
& 0.2601 & 0.3974 & 0.4198
& 0.3846 & 0.4872 & 0.5124 \\
Full CIM
& \textbf{0.2083} & \textbf{0.9231} & \textbf{0.9620}
& \textbf{0.2515} & \textbf{0.6154} & \textbf{0.6980}
& \textbf{0.1768} & \textbf{0.7949} & \textbf{0.7816} \\
\bottomrule
\end{tabular}
}
\vspace{-0.5em}
\end{table}

Table~\ref{tab:cim_quality} shows that Full CIM improves value-gap estimation across all three datasets. It reduces MAE from 0.4752 to 0.2083 on TOUCAN and from 0.3846 to 0.1768 on TRAJECT-Bench, while also improving Direction Accuracy and AUROC on the more challenging Toolathlon.
These results show that the full objective helps CIM estimate value gaps and identify the better tool choice.
Additional value-estimation metrics are provided in Appendix~\ref{app:cim_quality}.

Figure~\ref{fig:cim_calibration} further shows this trend. 
SFT and GRPO-only produce value-gap predictions with weak correlation to the ground truth. 
CIM trained with \(\mathcal{L}_{\text{value}}\) only reduces errors but still gives weak separation between positive and negative gaps. 
CITA shows the clearest alignment, supporting CIM as an effective comparative value estimator for tool decisions.

\begin{figure*}[t]
  \centering
  \includegraphics[width=\textwidth]{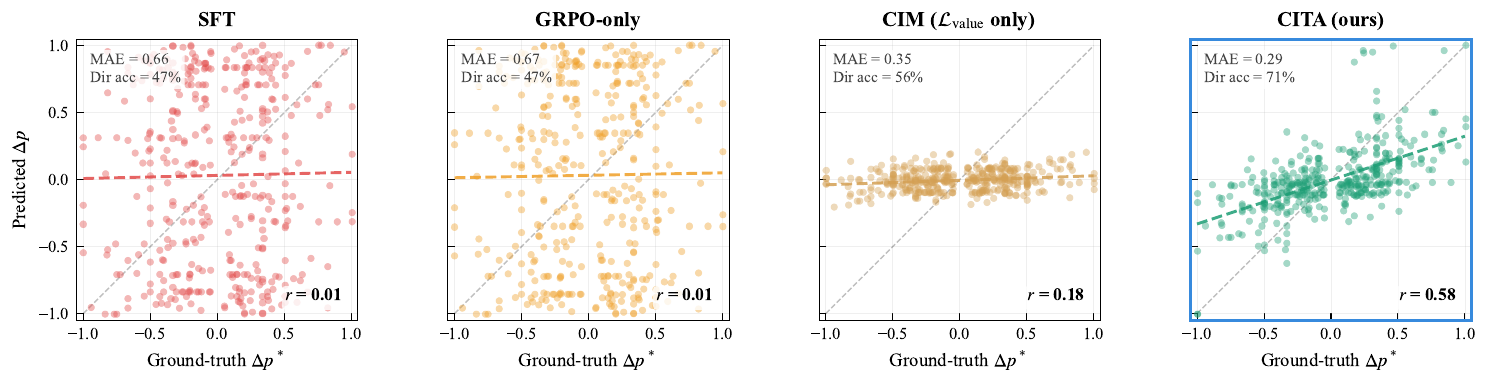}
  \caption{
  CIM value-gap estimation on held-out paired examples. Each point compares the predicted value gap \(\Delta\hat{v}\) with the observed value gap \(\Delta v\). Compared with SFT, GRPO-only, and CIM trained with \(\mathcal{L}_{\text{value}}\) only, CITA shows clearer alignment with the ground truth.
  }
  \label{fig:cim_calibration}
\end{figure*}

\subsection{Ablation Study}

\begin{wraptable}{r}{0.50\textwidth}
\vspace{-0.5em}
\centering
\small
\setlength{\tabcolsep}{4pt}
\renewcommand{\arraystretch}{1.05}
\caption{Ablation results averaged over six dataset-backbone settings.}
\label{tab:ablation}
\begin{tabular}{>{\centering\arraybackslash}p{0.46\linewidth}
                >{\centering\arraybackslash}p{0.22\linewidth}
                >{\centering\arraybackslash}p{0.18\linewidth}}
\toprule
Variant & Tool F1 & TSR \\
\midrule
\rowcolor{orange!8}
CITA (full) & \textbf{60.47} & \textbf{49.70} \\
w/o CIM guide & 55.12 & 43.68 \\
w/o \(r_{\mathrm{cmp}}\) & 56.21 & 44.37 \\
w/o \(\mathcal{L}_{\mathrm{gap}}\) & 54.86 & 41.92 \\
w/o confidence & 57.08 & 45.72 \\
w/o \(\mathcal{D}_{\mathrm{sandbox}}\) & 51.78 & 38.64 \\
w/o \(\mathcal{D}_{\mathrm{cmp}}\) & 55.78 & 43.88 \\
\bottomrule
\end{tabular}
\vspace{-0.6em}
\end{wraptable}

Table~\ref{tab:ablation} ablates the main components of CITA. 
Results are averaged over six dataset-backbone settings, covering three benchmarks and two backbone LLMs. 
Each metric is first computed within one setting and then averaged across settings. 
The variant w/o CIM guide removes CIM-guided candidate selection during rollout generation. 
The variant w/o \(r_{\mathrm{cmp}}\) removes the comparative reward and keeps only the task-level reward. 
The variant w/o \(\mathcal{L}_{\mathrm{gap}}\) trains CIM without explicit value-gap supervision. 
The variant w/o confidence uses \(V_\phi(s,a)\) alone for scoring and reward aggregation. 
The last two variants remove \(\mathcal{D}_{\mathrm{sandbox}}\) or \(\mathcal{D}_{\mathrm{cmp}}\) from CIM training.

Removing \(\mathcal{L}_{\mathrm{gap}}\) causes the largest loss-objective drop, showing that explicit value-gap learning is central to CIM. 
Removing CIM-guided rollout generation or \(r_{\mathrm{cmp}}\) also leads to clear degradation, supporting the dual role of CIM in trajectory construction and policy optimization. 
The confidence score gives a smaller but consistent gain, indicating that reliability-aware scoring helps reduce noisy tool choices. 
Among data-source ablations, removing \(\mathcal{D}_{\mathrm{sandbox}}\) causes the largest drop, confirming the importance of structured fork-point comparisons from the Bayesian simulator. 
Removing \(\mathcal{D}_{\mathrm{cmp}}\) also weakens performance, suggesting complementary benefits from real trajectory grounding and LLM-based semantic comparison. 
Detailed per-benchmark ablations are provided in Appendix~\ref{app:ablation}.

\section{Conclusion and Limitations}

This paper introduced CITA, a comparative inference framework for long-horizon tool use in LLM agents. CITA trains a Comparative Inference Model to compare candidate next-tool invocations under the current context and estimate their long-horizon value. The model is learned from paired signals built from real trajectories, a Bayesian tool-graph simulator, and LLM-based comparison, and is used to guide GRPO training and inference. Across three benchmarks, CITA improves Tool F1 and Task Success Rate. Value-estimation analysis and ablations further show that CIM learns meaningful value gaps. The current study focuses on benchmarked tool environments with available logged trajectories and tool graphs, which provides a controlled setting for evaluating comparative value estimation. Extending this framework to more dynamic tool ecosystems, including settings where tools are added, removed, or updated over time, is a natural next step. 

\bibliographystyle{unsrtnat}
\bibliography{ref}

\newpage


\appendix

\section{Dataset Statistics}
\label{app:dataset_stats}

We evaluate CITA on three tool-use benchmarks with different sources of difficulty. Toolathlon is designed for diverse, realistic, and long-horizon task execution across software environments. TOUCAN provides large-scale tool-agentic trajectories synthesized from real-world MCP environments. TRAJECT-Bench focuses on trajectory-aware tool-use evaluation, where the quality of tool selection, parameterization, and ordering is evaluated beyond the final answer. These benchmarks allow us to test whether CITA improves tool-use agents under long trajectories, large tool pools, and trajectory-level constraints.

\begin{table}[ht]
\centering
\scriptsize
\caption{Statistics of the processed benchmark splits used in our experiments.}
\label{tab:dataset_stats}
\begin{tabular}{lccc}
\toprule
Benchmark & Avg. tool invocations / episode & \# Unique tools & Main difficulty \\
\midrule
Toolathlon & \(\sim 13.5\) & 1,250 & Long horizon and multi-step \\
TOUCAN & \(\sim 3.2\) & 15,294 & Large tool pool \\
TRAJECT-Bench & \(\sim 6.4\) & 715 & Medium horizon and trajectory quality \\
\bottomrule
\end{tabular}
\end{table}

Toolathlon has the longest trajectories in our evaluation, with about 13.5 tool invocations per episode. It is therefore well suited for testing whether an agent can maintain useful tool choices across many dependent steps. TOUCAN has much shorter trajectories, with about 3.2 calls per episode, but its tool pool is far larger than the other two benchmarks. It tests whether the agent can select appropriate tools from a broad set of possible APIs. TRAJECT-Bench lies between these two settings, with about 6.4 calls per episode and 715 tools. Its trajectory-aware design makes it useful for evaluating whether the generated tool-use path is coherent, not only whether the final answer is correct.

The three benchmarks are complementary. Toolathlon stresses long-horizon dependency, TOUCAN stresses large-scale tool selection, and TRAJECT-Bench stresses medium-horizon trajectory quality. This combination allows us to evaluate CITA across different types of tool-use difficulty rather than a single benchmark setting.

\section{Training Dynamics}
\label{app:reward_curves}

\begin{figure*}[ht]
  \centering
  \includegraphics[width=\textwidth]{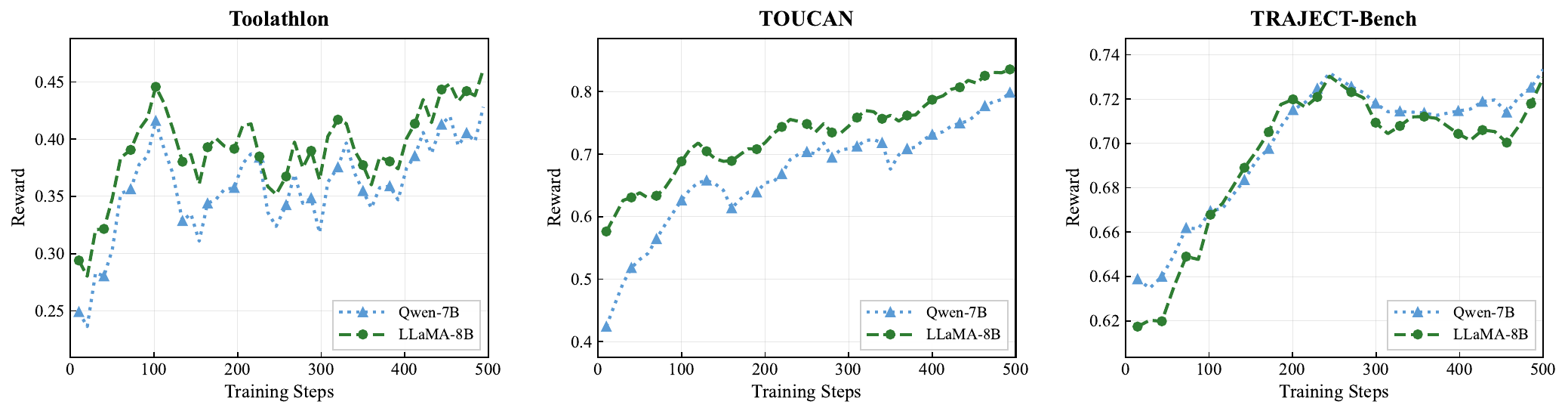}
  \caption{
  Training reward curves during CIM-guided GRPO. Rewards generally increase over training steps on Toolathlon, TOUCAN, and TRAJECT-Bench for both Qwen2.5-7B and Llama3.1-8B, showing that CITA provides useful optimization signals across different tool-use settings.
  }
  \label{fig:reward_curves_app}
\end{figure*}

Figure~\ref{fig:reward_curves_app} reports the training reward curves of CITA during CIM-guided GRPO on Toolathlon, TOUCAN, and TRAJECT-Bench. The curves are plotted for Qwen2.5-7B and Llama3.1-8B under the same training setting. These results provide additional evidence that CIM-guided rollouts and comparative rewards give useful optimization signals during policy learning.

The three benchmarks show different training patterns. On Toolathlon, rewards increase overall but fluctuate more strongly than on the other two datasets. This is consistent with the long-horizon nature of Toolathlon, where each episode contains more tool invocations and early decisions can affect many later steps. In this setting, a small change in tool selection can lead to a large change in trajectory quality, which makes the reward curve less smooth. Even with these fluctuations, both backbones show a clear upward trend during training.

TOUCAN shows the most stable improvement. Although its trajectories are shorter, the benchmark contains a very large tool pool, which makes next-tool selection difficult. The reward curves increase steadily for both backbones, suggesting that CIM-guided selection helps the policy identify useful tools from a broad action space. This trend is consistent with the main results in Table~\ref{tab:main_results}, where CITA obtains its largest gains on TOUCAN.

TRAJECT-Bench shows a different pattern. Rewards increase quickly in the early stage and then enter a relatively stable range. This suggests that the policy first learns to generate more coherent tool-use trajectories and then makes smaller refinements. Since TRAJECT-Bench emphasizes trajectory-level quality, the plateau is expected once the agent has learned a reliable tool-use pattern. CITA still maintains strong final performance on this benchmark, which indicates that comparative value estimation remains useful even when planning-based baselines are already competitive.

Overall, the reward curves support the main experimental finding. CITA does not only improve final evaluation metrics after training; it also provides a stable optimization signal during GRPO. The upward trends across all three benchmarks indicate that CIM-guided rollouts and comparative rewards can guide policy learning under long-horizon dependency, large tool spaces, and medium-horizon trajectory evaluation.

\section{Additional CIM Value Estimation Results}
\label{app:cim_quality}

This section provides additional details for the CIM value estimation evaluation. 
The goal is to test whether CIM can predict the value difference between two next-tool choices under the same or comparable context. 
For each held-out pair, CIM predicts \(\Delta\hat{v}=\hat{v}_f-\hat{v}_c\), and we compare it with the observed value gap \(\Delta v\). 
The evaluation set is built from observed trajectory pairs with sufficient support. 
Following the evaluation protocol, we separate strong-effect pairs and near-neutral pairs by the magnitude of the observed value gap. 
Strong-effect pairs satisfy \(|\Delta v|\ge 0.15\), while near-neutral pairs satisfy \(|\Delta v|\le 0.05\). 
This split allows us to evaluate not only numerical value-gap estimation, but also whether CIM can distinguish meaningful tool-choice differences from weak or negligible differences.

\paragraph{Core metrics.}
The main text reports three metrics. 
\textbf{Value Gap MAE} measures the mean absolute error between the predicted and observed value gaps, \(|\Delta\hat{v}-\Delta v|\). 
It evaluates whether CIM estimates the size of the value difference correctly. 
\textbf{Direction Accuracy} measures whether \(\Delta\hat{v}\) and \(\Delta v\) have the same sign. 
It evaluates whether CIM identifies which tool choice has the higher long-horizon value. 
This metric is directly related to decision making, because the agent usually needs to choose the better tool rather than predict the exact value gap. 
\textbf{AUROC} evaluates whether the magnitude \(|\Delta\hat{v}|\) can distinguish strong-effect pairs from near-neutral pairs. 
It measures whether CIM assigns larger predicted gaps to pairs where the two tool choices truly differ.

\paragraph{Additional discrimination metrics.}
We also report \textbf{AUPRC} and \textbf{CvS}. 
AUPRC uses the same score \(|\Delta\hat{v}|\) as AUROC, but focuses on the precision-recall tradeoff for identifying strong-effect pairs. 
It is useful when the number of strong-effect and near-neutral pairs is imbalanced. 
CvS, short for contrast-vs-small-gap score, measures whether strong-effect pairs receive predicted gaps above a fixed threshold and near-neutral pairs receive predicted gaps below the threshold.
In our evaluation, the threshold is \(0.1\). 
A high CvS means that CIM not only detects meaningful differences, but also avoids assigning large gaps to near-neutral comparisons.

\paragraph{Correlation metrics.}
\textbf{Pearson} and \textbf{Spearman} correlations measure agreement between predicted and observed value gaps from two complementary views. 
Pearson \(r\) measures linear agreement between \(\Delta\hat{v}\) and \(\Delta v\). 
It is high when the predicted gap increases proportionally with the observed gap. 
Spearman \(r\) measures rank-level agreement. 
It is high when CIM orders pairs by value-gap strength in a way that matches the observed ordering, even if the exact scale is not perfectly calibrated.


\begin{table}[ht]
\centering
\small
\setlength{\tabcolsep}{5pt}
\renewcommand{\arraystretch}{1.08}
\caption{Additional CIM value estimation metrics. CIM w/ \(\mathcal{L}_{\text{value}}\) only uses only the value prediction loss, while Full CIM uses the complete objective with value-gap learning, consistency regularization, and confidence calibration.}
\label{tab:cim_quality_appendix}
\begin{tabular}{llcccc}
\toprule
Dataset & Model & AUPRC \(\uparrow\) & CvS \(\uparrow\) & Pearson \(r\) \(\uparrow\) & Spearman \(r\) \(\uparrow\) \\
\midrule
\multirow{2}{*}{TOUCAN}
& CIM w/ \(\mathcal{L}_{\text{value}}\) only & 0.9945 & 0.2667 & 0.7652 & \textbf{0.8965} \\
& Full CIM & \textbf{0.9945} & \textbf{0.9333} & \textbf{0.9004} & 0.8700 \\
\midrule
\multirow{2}{*}{Toolathlon}
& CIM w/ \(\mathcal{L}_{\text{value}}\) only & 0.5401 & 0.4091 & -0.1307 & -0.1884 \\
& Full CIM & \textbf{0.7560} & \textbf{0.5303} & \textbf{0.3077} & \textbf{0.2193} \\
\bottomrule
\end{tabular}
\end{table}

The additional metrics support the conclusion in the main text \ref{tab:cim_quality}. 
On TOUCAN, Full CIM greatly improves CvS from 0.2667 to 0.9333 and increases Pearson correlation from 0.7652 to 0.9004, indicating that the complete objective produces value gaps that are better calibrated and more discriminative. 
AUPRC is already high for CIM w/ \(\mathcal{L}_{\text{value}}\) only and remains high for Full CIM, which suggests that this dataset is already separable under the AUPRC metric. 
Although Spearman correlation slightly decreases from 0.8965 to 0.8700, it remains high, while the large gains in CvS and Pearson correlation show that Full CIM better separates true value differences from near-neutral comparisons.

Toolathlon is more difficult because its trajectories are longer and the effect of one tool choice may depend on many later decisions. 
Under this setting, CIM w/ \(\mathcal{L}_{\text{value}}\) only has weak agreement with the observed gaps, with negative Pearson and Spearman correlations. 
Full CIM improves all reported metrics on Toolathlon, raising AUPRC from 0.5401 to 0.7560, CvS from 0.4091 to 0.5303, Pearson correlation from -0.1307 to 0.3077, and Spearman correlation from -0.1884 to 0.2193. 
This improvement suggests that the complete CIM objective helps the model recover meaningful value-gap structure when long-horizon dependencies make the signal noisier.

\section{Additional Ablation Results}
\label{app:ablation}

This section provides detailed ablation results for CITA. The main table \ref{tab:ablation} reports averaged results over six dataset-backbone settings . Here we expand the results for each benchmark and backbone, and include additional ablation variants.

\paragraph{Ablation variants.}
We consider the following variants. 
\begin{itemize}
    \item \textit{w/o CIM guide} removes CIM-guided candidate selection during rollout generation.
    \item \textit{w/o \(r_{\mathrm{cmp}}\)} removes the comparative reward and uses only the task-level reward.
    \item \textit{w/o \(\mathcal{L}_{\mathrm{gap}}\)} trains CIM without explicit value-gap supervision.
    \item \textit{w/o \(\mathcal{L}_{\mathrm{cons}}\)} removes the consistency regularization on value gaps.
    \item \textit{w/o confidence} uses \(V_\phi(s,a)\) alone for candidate scoring and reward aggregation.
    \item \textit{w/o \(\mathcal{D}_{\mathrm{obs}}\)}, \textit{w/o \(\mathcal{D}_{\mathrm{sandbox}}\)}, and \textit{w/o \(\mathcal{D}_{\mathrm{cmp}}\)} remove the corresponding training source for CIM.
    \item \textit{Uniform mixing} removes phase-aware data mixing and uses a fixed source mixture during CIM training.
    \item \textit\textit{\(K=1\) candidate} keeps CIM in the rollout process but samples only one candidate tool invocation at each step, so CIM cannot compare multiple alternatives under the same context.
    \item \textit{Mean aggregation} replaces confidence-weighted reward aggregation with a simple mean over step values.
\end{itemize}

Results are reported in Tool F1 and TSR. The last two columns show the arithmetic average over the six dataset-backbone settings.

\begin{table*}[t]
\centering
\scriptsize
\setlength{\tabcolsep}{2.4pt}
\renewcommand{\arraystretch}{1.06}
\caption{Detailed ablation results on each dataset-backbone setting. Avg. is the arithmetic mean over all six settings.}
\label{tab:ablation_full}
\resizebox{\textwidth}{!}{
\begin{tabular}{lcccccccccccccc}
\toprule
\multirow{2}{*}{Variant}
& \multicolumn{2}{c}{Toolathlon-Qwen}
& \multicolumn{2}{c}{Toolathlon-Llama}
& \multicolumn{2}{c}{TOUCAN-Qwen}
& \multicolumn{2}{c}{TOUCAN-Llama}
& \multicolumn{2}{c}{TRAJECT-Qwen}
& \multicolumn{2}{c}{TRAJECT-Llama}
& \multicolumn{2}{c}{Avg.} \\
\cmidrule(lr){2-3}\cmidrule(lr){4-5}\cmidrule(lr){6-7}\cmidrule(lr){8-9}\cmidrule(lr){10-11}\cmidrule(lr){12-13}\cmidrule(lr){14-15}
& F1 & TSR & F1 & TSR & F1 & TSR & F1 & TSR & F1 & TSR & F1 & TSR & F1 & TSR \\
\midrule
CITA (full)
& \textbf{58.56} & \textbf{47.82}
& \textbf{59.29} & \textbf{45.05}
& \textbf{59.67} & \textbf{54.23}
& \textbf{56.44} & \textbf{47.08}
& \textbf{59.30} & \textbf{48.31}
& \textbf{69.58} & \textbf{55.72}
& \textbf{60.47} & \textbf{49.70} \\
\midrule
w/o CIM guide
& 53.21 & 39.62
& 54.38 & 38.97
& 55.40 & 48.10
& 52.16 & 40.55
& 54.09 & 41.23
& 61.48 & 53.61
& 55.12 & 43.68 \\
w/o \(r_{\mathrm{cmp}}\)
& 54.06 & 40.88
& 55.73 & 39.76
& 56.82 & 48.62
& 53.39 & 41.31
& 55.84 & 42.05
& 61.42 & 53.60
& 56.21 & 44.37 \\
w/o \(\mathcal{L}_{\mathrm{gap}}\)
& 52.94 & 37.91
& 53.28 & 36.85
& 55.11 & 47.06
& 51.72 & 38.77
& 53.68 & 39.65
& 62.43 & 51.28
& 54.86 & 41.92 \\
w/o \(\mathcal{L}_{\mathrm{cons}}\)
& 56.21 & 43.20
& 57.46 & 41.92
& 58.20 & 50.61
& 55.30 & 44.08
& 56.94 & 45.60
& 63.89 & 54.37
& 58.00 & 46.63 \\
w/o confidence
& 55.14 & 42.15
& 56.21 & 40.88
& 57.43 & 49.36
& 54.02 & 42.71
& 56.18 & 43.82
& 63.50 & 55.40
& 57.08 & 45.72 \\
\midrule
w/o \(\mathcal{D}_{\mathrm{obs}}\)
& 50.43 & 35.60
& 50.91 & 34.28
& 54.05 & 46.12
& 49.38 & 36.50
& 51.24 & 38.44
& 58.21 & 46.54
& 52.37 & 39.58 \\
w/o \(\mathcal{D}_{\mathrm{sandbox}}\)
& 50.10 & 34.82
& 50.62 & 33.97
& 53.18 & 44.28
& 49.06 & 35.74
& 51.02 & 37.65
& 56.72 & 45.40
& 51.78 & 38.64 \\
w/o \(\mathcal{D}_{\mathrm{c m p}}\)
& 53.70 & 39.84
& 54.80 & 38.65
& 55.31 & 48.03
& 52.67 & 40.22
& 55.03 & 42.15
& 63.17 & 54.39
& 55.78 & 43.88 \\
Uniform mixing
& 55.67 & 42.38
& 56.55 & 41.75
& 57.91 & 50.26
& 53.96 & 43.65
& 56.82 & 44.29
& 62.29 & 54.87
& 57.20 & 46.20 \\
\(K=1\) candidate
& 52.88 & 38.75
& 53.76 & 37.90
& 55.50 & 47.25
& 51.88 & 39.16
& 54.34 & 40.73
& 61.04 & 52.77
& 54.90 & 42.76 \\
Mean aggregation
& 56.42 & 43.06
& 57.08 & 41.48
& 58.62 & 50.71
& 54.85 & 44.19
& 57.15 & 45.18
& 65.86 & 54.62
& 58.33 & 46.54 \\
\bottomrule
\end{tabular}
}
\end{table*}

\paragraph{Analysis.}
The detailed results support the averaged findings in the main table \ref{tab:ablation}. Removing \(\mathcal{L}_{\mathrm{gap}}\) gives the largest drop among the main variants, reducing the average Tool F1 from 60.47 to 54.86 and TSR from 49.70 to 41.92. This confirms that explicit supervision on value gaps is central to CIM, because the model must learn not only whether one tool invocation is useful, but also whether it is better than another choice under the same context.

The two uses of CIM are both important. Removing CIM-guided rollout generation lowers the average Tool F1 to 55.12 and TSR to 43.68, while removing the comparative reward lowers them to 56.21 and 44.37. The first result shows that CIM helps construct better trajectories by selecting among candidate tool invocations before execution. The second shows that CIM also provides useful training feedback after trajectories are generated. This supports the dual role of CIM in CITA.

The data-source ablations show that the three sources provide complementary supervision. Removing \(\mathcal{D}_{\mathrm{sandbox}}\) causes the largest drop among the data-source variants, reducing the average Tool F1 from 60.47 to 51.78 and TSR from 49.70 to 38.64. This confirms the importance of Bayesian simulator pairs, which provide scalable and structured fork-point comparisons that are difficult to obtain from logged trajectories alone. Removing \(\mathcal{D}_{\mathrm{obs}}\) also causes a clear drop, showing that observed trajectory pairs are needed to anchor CIM to real tool-use behavior. Removing \(\mathcal{D}_{\mathrm{cmp}}\) further weakens performance, indicating that LLM-based comparative comparison supplies useful semantic signals for rare or unseen alternatives.

The additional variants further clarify the design choices. Uniform mixing underperforms the full phase-aware mixing schedule, suggesting that early structured supervision and later real-distribution calibration are both useful. The \(K=1\) candidate variant performs poorly because CIM cannot compare alternatives when the policy provides only one candidate at each step. Mean aggregation is better than removing confidence entirely, but still worse than full CITA, which shows that confidence-weighted aggregation helps reduce the influence of uncertain value estimates.

Across datasets, the largest drops often appear on Toolathlon and TOUCAN. Toolathlon is sensitive to early tool-choice errors because of its long trajectories, while TOUCAN is sensitive to candidate selection because of its large tool pool. TRAJECT-Bench is more forgiving in some variants, especially under Llama3.1-8B, but full CITA still gives the best overall performance. These patterns suggest that the components of CITA are useful under different tool-use difficulties rather than only one benchmark setting.

\section{Implementation Details}
\label{app:implementation}

This section provides implementation details for the experiments. Additional details on paired data construction are provided in Appendix~\ref{app:data_construction}. Unless otherwise specified, all experiments use the same train-test split, tool environment, and evaluation protocol across methods.

\paragraph{Backbone models.}
We use Qwen2.5-7B and Llama3.1-8B as backbone LLMs. Both models are first adapted to the tool-use format with supervised fine-tuning and then optimized with the corresponding training method. The same tokenizer, tool registry, simulator, and evaluation scripts are used for all methods.

\paragraph{Tool-use format.}
All models follow a unified tool invocation format. The system prompt lists the available tools for the current task and requires the model to emit a structured invocation:
\[
\texttt{<tool\_call>\{"name": "...", "arguments": \{...\}\}</tool\_call>}.
\]
After each invocation, the tool response is appended to the dialogue context, and the model either invokes another tool or produces a final response. For each task, we include tools from successful reference trajectories, add related tools, and fill the remaining slots with randomly sampled tools. We use at most 50 tools in the prompt.

\paragraph{Supervised fine-tuning.}
SFT data are constructed from successful tool-use episodes. Each example is formatted as a multi-turn dialogue containing the user request, tool invocations, tool responses, and the final response. We also include continuation examples that start from intermediate trajectory prefixes, exposing the model to partially completed tool-use states. SFT is performed with LoRA and assistant-only loss masking, where loss is computed only on assistant outputs. For the 7B and 8B backbones, we use LoRA rank 64, LoRA alpha 128, learning rate \(5\times10^{-6}\), maximum sequence length 4096, and train for 3 epochs.

\paragraph{CIM training.}
CIM is trained after SFT and before GRPO. The model uses the LLM backbone as a semantic encoder and projects hidden states into a lower-dimensional value-estimation space. In our main configuration, the CIM hidden dimension is 768, and the CIM module uses 8 Transformer layers with 12 attention heads. The final decision representation is passed to two lightweight MLP heads: a value head for \(V_\phi(s,a)\) and a confidence head for \(c_\phi(s,a)\). The training objective follows Eq.~\ref{eq:cim_loss}. We train CIM with a progressive schedule that gradually increases the difficulty of supervision. In the first stage, we activate only \(\mathcal{L}_{\mathrm{value}}\), so the model learns stable estimates of \(V_\phi(s,a)\) before being asked to compare alternatives. In the second stage, we add \(\mathcal{L}_{\mathrm{gap}}\) and \(\mathcal{L}_{\mathrm{cons}}\), which trains CIM to match value differences between paired tool choices and regularizes the comparison space with self-identity and transitivity constraints. In the final stage, we add \(\mathcal{L}_{\mathrm{conf}}\) to calibrate the confidence head after the value estimates have become stable. This schedule avoids forcing value prediction, pairwise comparison, and confidence calibration to be learned from scratch at the same time. For CIM training, we use batch size 8, gradient accumulation 8, backbone learning rate \(1\times10^{-5}\), CIM learning rate \(5\times10^{-5}\), and AdamW optimization with a warmup-cosine learning-rate schedule.

\paragraph{GRPO training.}
For CITA, we initialize the policy from the SFT model and optimize it with CIM-guided GRPO. The SFT LoRA adapter is merged into the base model, and a new trainable actor LoRA adapter is attached for GRPO. The trained CIM is kept frozen during policy optimization and is used to score candidate tool invocations and compute comparative rewards. We use LoRA rank 64, LoRA alpha 128, learning rate \(2\times10^{-6}\), group size 6, gradient accumulation 2, sampling temperature 0.7, and maximum rollout length 15. The reward is the weighted combination in Eq.~\ref{eq:total_reward}, with \(w_{\mathrm{task}}=0.4\) and \(w_{\mathrm{cmp}}=0.6\).

\paragraph{CIM-guided rollout and inference.}
During GRPO rollout generation, the policy samples multiple candidate tool invocations at each decision step. In the main setting, we sample 3 candidates per step. CIM scores each candidate using Eq.~\ref{eq:cim_score}, and the highest-scoring invocation is executed. The returned observation is appended to the context before the next step. Inference uses the same CIM-guided selection rule as rollout generation, but no policy update is performed.

\paragraph{Data split and reproducibility.}
We use a deterministic task-level split with seed 42. Test tasks are excluded from SFT continuation construction and GRPO prompt sampling. For datasets where task names are not unique enough to define tasks, we use the user prompt as the task key. All reported results use the same held-out split across methods.

\paragraph{Hardware and precision.}
We train 7B and 8B models with mixed precision and gradient checkpointing. SFT uses HuggingFace Trainer and DeepSpeed ZeRO-2 when needed. GRPO uses a custom training loop with a shared base LLM, a trainable actor LoRA adapter, and a frozen CIM module. Main experiments are run on 8x H100 GPUs.

\section{Data Construction Details}
\label{app:data_construction}

This section provides additional details on how we construct paired training signals for CIM. All sources are converted into the paired format used in Eq.~\ref{eq:paired_signal}:
\begin{equation}
x=(s,a_f,a_c,v_f,v_c,\Delta v),
\qquad
\Delta v=v_f-v_c .
\end{equation}
Here \(s\) is the tool-use context, \(a_f\) and \(a_c\) are two alternative next-tool invocations, and \(v_f\) and \(v_c\) are their estimated long-horizon success values. In implementation, we also store an optional sample weight \(w\) and source-specific metadata, such as support counts, sandbox paths, and LLM confidence scores. The three sources differ in how they obtain the paired choices and their value labels.

\subsection{\(\mathcal{D}_{\mathrm{obs}}\): Observed Trajectory Pairs}

\(\mathcal{D}_{\mathrm{obs}}\) is mined from logged tool-use trajectories. This source keeps part of CIM training close to real agent behavior. For each episode, we record the task name, tool sequence, tool arguments, tool outputs, and trajectory success label. At each step, we build a context key that includes three views of the prefix: the prefix tool sequence, compact hashes of previous arguments, and signatures of recent tool outputs. This richer context key avoids grouping together prefixes that use the same tools but have different arguments or observations.

For each task and context \(s\), we collect all next-tool invocations observed after that context. For a candidate next tool \(a\), we estimate its empirical value by the downstream success rate:
\begin{equation}
\hat{v}_{\mathrm{obs}}(s,a)=
\frac{N_{\mathrm{succ}}(s,a)}{N(s,a)}.
\end{equation}
When two different tools \(a_f\) and \(a_c\) are observed under the same context, we construct a paired example by setting
\begin{equation}
v_f=\hat{v}_{\mathrm{obs}}(s,a_f),
\qquad
v_c=\hat{v}_{\mathrm{obs}}(s,a_c),
\qquad
\Delta v=v_f-v_c .
\end{equation}
We require a minimum support for each context-action estimate and limit the number of pairs per context to avoid overrepresenting frequent contexts. The sample weight is based on the available support, giving higher weight to comparisons backed by more observations. In addition to the tool prefix, we keep compact argument summaries and output summaries, so the text encoder can recover state information during CIM training.

\subsection{\texorpdfstring{$\mathcal{D}_{\mathrm{sandbox}}$}{D	extsubscript{sandbox}}: Bayesian Simulator Pairs}

\(\mathcal{D}_{\mathrm{sandbox}}\) is generated from the tool transition graph. This source provides scalable and structured comparisons at fork points, especially when logged trajectories do not cover enough alternatives under the same context.

We first build a directed tool transition graph from logged trajectories. Each node denotes a tool, and each directed edge \(i\rightarrow j\) records that tool \(j\) was invoked after tool \(i\). A fork point is a node with at least two outgoing successors. For each sandbox instance, we sample a fixed set of synthetic probabilities for all tool nodes and transition edges. For a tool node \(i\), the sampled node success probability is
\begin{equation}
p_i \sim \mathrm{Beta}(g n_i+\lambda,(1-g)n_i+\lambda),
\end{equation}
where \(n_i\) is the tool invocation count, \(g\) is the global success rate in the logged data, and \(\lambda\) is a smoothing constant. For an edge \(i\rightarrow j\), the transition reliability is sampled from a Beta prior parameterized by the observed transition count and the relative frequency of this edge among outgoing transitions from \(i\):
\begin{equation}
r_{ij}\sim\mathrm{Beta}(q_{ij}n_{ij}+\lambda,(1-q_{ij})n_{ij}+\lambda),
\qquad
q_{ij}=\frac{n_{ij}}{\sum_{j'}n_{ij'}} .
\end{equation}
All comparisons inside the same sandbox share the sampled \(p_i\) and \(r_{ij}\), forming a fixed virtual world.

For each fork node \(u\), we generate a plausible prefix context by a reverse walk on the tool graph. The resulting prefix ends at \(u\) and serves as the context \(s\). We then choose two successor tools \(a_f\) and \(a_c\) from the outgoing neighbors of \(u\). Starting from each successor, we expand a short continuation path inside the same sandbox. For a path \(P=(b_0,\ldots,b_L)\), the sandbox value is computed as
\begin{equation}
\hat{v}_{\mathrm{sandbox}}(P)
=
\prod_{\ell=0}^{L}p_{b_\ell}
\prod_{\ell=0}^{L-1}r_{b_\ell b_{\ell+1}} .
\end{equation}
The two expanded paths \(P_f\) and \(P_c\) define
\begin{equation}
v_f=\hat{v}_{\mathrm{sandbox}}(P_f),
\qquad
v_c=\hat{v}_{\mathrm{sandbox}}(P_c),
\qquad
\Delta v=v_f-v_c .
\end{equation}
Since both paths are evaluated in the same sampled sandbox, the comparison isolates the effect of choosing different successor tools while keeping the synthetic environment fixed.

In implementation, we generate multiple sandbox instances with different sampled probabilities. Each sandbox enumerates fork points and successor pairs, with limits on the number of sampled successors and the continuation depth to control data size. This source is especially useful for training value-gap consistency because multiple comparisons inside the same sandbox share the same underlying node and edge probabilities.

\subsection{\(\mathcal{D}_{\mathrm{cmp}}\): LLM-Based comparative Pairs}

\(\mathcal{D}_{\mathrm{cmp}}\) provides semantic comparisons for alternatives that are rare in logged trajectories or difficult to cover with the graph simulator. For each logged trajectory, we select a prefix \(s\), keep the factual next tool \(a_f\), and choose a plausible comparative tool \(a_c\). The comparative tool is selected by a three-level strategy. We first prefer other valid successors of the current predecessor in the tool graph, because these alternatives are position-sensitive. At the first step, where there may be no predecessor, we use semantically related entry tools. If neither source provides a valid candidate, we use tools from the same server or tool family as a fallback.

The factual and comparative choices are evaluated in the same LLM prompt. Given the task description and the executed prefix state, the LLM is asked to imagine a short continuation for each option, including subsequent tool invocations and state changes, and then judge whether each branch can complete the task. Using a paired prompt is important because both options are judged under the same context and the same calibration scale.

We repeat the paired comparison \(M\) times. In the main setting, \(M=5\). Only valid parsed responses are kept. Let \(M_s\) be the number of valid comparisons. For the \(m\)-th valid comparison, \(y_f^{(m)}\in\{0,1\}\) and \(y_c^{(m)}\in\{0,1\}\) denote whether the LLM judges the continuations after \(a_f\) and \(a_c\) as successful. The LLM-based value estimates are
\begin{equation}
\hat{v}_{\mathrm{cmp}}(s,a_f)=
\frac{1}{M_s}\sum_{m=1}^{M_s}y_f^{(m)},
\qquad
\hat{v}_{\mathrm{cmp}}(s,a_c)=
\frac{1}{M_s}\sum_{m=1}^{M_s}y_c^{(m)} .
\end{equation}
We retain a pair only if enough valid responses are parsed. In our implementation, at least three valid comparisons are required. We also compute a consistency score from the repeated judgments and combine it with parsing coverage to form a confidence score. Low-confidence pairs are filtered out, and the remaining pair weight is proportional to the confidence. This filtering reduces noise from unstable LLM judgments while preserving semantic comparisons that are difficult to obtain from logs alone.

\paragraph{LLM-based comparison prompt.}
The prompt used for LLM-based comparison is shown below. 
The prompt places the factual and comparative tool choices in the same context and asks the LLM to imagine one plausible continuation for each option. 
This paired format reduces calibration drift between separate LLM calls and makes the two success judgments more directly comparable.

\begin{verbatim}
You are simulating possible future trajectories for an AI agent's 
tool selection.

Task: {task_description}

Executed prefix state:
{context_state}

The agent must choose between two tools. For EACH option, imagine 
one plausible
continuation trajectory (1-5 tool invocations, until task success 
or clear failure).
For each step: tool name, one-line state change.

Option A: {factual_tool_name} -- {factual_tool_desc}
Option B: {counter_tool_name} -- {counter_tool_desc}

Respond in JSON:
{
  "trajectory_a": [
    {"tool": "<name>", "state_change": "<brief>"}
  ],
  "success_a": <0_or_1>,
  "trajectory_b": [
    {"tool": "<name>", "state_change": "<brief>"}
  ],
  "success_b": <0_or_1>,
  "reasoning": "<one sentence>"
}
\end{verbatim}

\subsection{Phase-Aware Dataset Mixing}

After constructing the three sources, we normalize them into the same pair format and merge them for CIM training. We keep source-specific fields, such as support counts, sandbox paths, and LLM confidence, because they can be useful for weighting and diagnostics. For implementation, we instantiate the progressive mixture with three training phases, summarized in Table~\ref{tab:phase_mixing}.

\begin{table}[ht]
\centering
\small
\setlength{\tabcolsep}{5pt}
\renewcommand{\arraystretch}{1.08}
\caption{Phase-aware mixing of the three paired data sources. Source weights affect the training weight of examples, while sample ratios affect how many examples are sampled from each source.}
\label{tab:phase_mixing}
\begin{tabular}{lcccccc}
\toprule
\multirow{2}{*}{Phase}
& \multicolumn{3}{c}{Source weight}
& \multicolumn{3}{c}{Sample ratio} \\
\cmidrule(lr){2-4}\cmidrule(lr){5-7}
& \(\mathcal{D}_{\mathrm{obs}}\) & \(\mathcal{D}_{\mathrm{sandbox}}\) & \(\mathcal{D}_{\mathrm{cmp}}\)
& \(\mathcal{D}_{\mathrm{obs}}\) & \(\mathcal{D}_{\mathrm{sandbox}}\) & \(\mathcal{D}_{\mathrm{cmp}}\) \\
\midrule
Phase 1 & 0.8 & 1.4 & 0.0 & 0.6 & 1.0 & 0.0 \\
Phase 2 & 1.0 & 1.0 & 1.0 & 0.8 & 1.0 & 1.0 \\
Phase 3 & 1.8 & 0.3 & 1.2 & 1.0 & 0.3 & 1.0 \\
\bottomrule
\end{tabular}
\end{table}

The phases follow the same motivation as the progressive CIM training schedule. Phase 1 emphasizes \(\mathcal{D}_{\mathrm{sandbox}}\), which provides clean structural supervision and many controlled fork-point comparisons. Phase 2 balances all three sources, allowing the model to learn value prediction, value-gap estimation, and semantic comparison jointly. Phase 3 increases the influence of \(\mathcal{D}_{\mathrm{obs}}\) and \(\mathcal{D}_{\mathrm{cmp}}\), which calibrates CIM toward real tool-use contexts while still retaining a small amount of simulator supervision.

\subsection{Train-Validation Split and Leakage Control}

We use a group-aware split when creating CIM train and validation sets. The split key includes the source, task identifier, context prefix, argument-pattern hashes, and recent output signatures. For sandbox pairs, the split key also includes the sandbox identifier and context. We further group mirror comparisons by sorting the factual and comparative action identifiers. This prevents a pair \((a_f,a_c)\) from appearing in training while its mirrored comparison \((a_c,a_f)\) appears in validation under the same context.

This split is important because paired comparisons often come in symmetric or closely related forms. A naive random split could leak the same state family into both train and validation sets, leading to overestimated value-estimation quality. The group-aware split keeps related comparisons in the same split and provides a more reliable validation signal for CIM.

\section{Evaluation Protocol}
\label{app:evaluation_protocol}

This section describes the general evaluation protocol used in our experiments. The GPT-4o judge protocol for Task Success Rate is provided separately in Appendix~\ref{app:gpt4o_judge}. All methods are evaluated on the same held-out split, with the same tool registry, tool simulator, maximum step budget, and evaluation prompts.

\paragraph{Held-out split.}
We use a deterministic task-level split with seed 42. For datasets with meaningful task names, the task name is used as the split key. For datasets where task names are not sufficiently discriminative, we use the user prompt as the task key. We assign 20\% of task keys to the test split by hashing the task key with the seed. Test tasks are excluded from GRPO prompt sampling and training-time trajectory construction.

\paragraph{Trajectory collection.}
For each test task, the agent interacts with the same tool simulator. At each step, the model emits a structured tool invocation:
\[
\texttt{<tool\_call>\{"name": "...", "arguments": \{...\}\}</tool\_call>}.
\]
The simulator returns the corresponding observation, which is appended to the dialogue context. The interaction terminates when the model produces a final response or reaches the maximum step limit. We record the user request, all tool invocations, tool arguments, tool observations, and the final response.

\paragraph{Tool-name normalization.}
Before computing tool-matching metrics, we normalize tool names by removing version suffixes such as \(\texttt{\_v1}\) or \(\texttt{\_v13}\). We also ignore minor formatting differences such as case, hyphen, and underscore variations when the normalized tool names indicate the same function.

\paragraph{Tool F1.}
Tool F1 evaluates whether the generated trajectory selects tools that match successful reference trajectories. Let \(P=(p_1,\ldots,p_m)\) be the generated tool sequence and \(R=(r_1,\ldots,r_n)\) be a reference sequence. We compute an order-preserving fuzzy match between normalized tool names and denote the number of matched invocations by \(M(P,R)\). Precision, recall, and F1 are computed as
\begin{equation}
\mathrm{Precision}=\frac{M(P,R)}{|P|}, \qquad
\mathrm{Recall}=\frac{M(P,R)}{|R|}, \qquad
\mathrm{F1}=\frac{2\cdot\mathrm{Precision}\cdot\mathrm{Recall}}{\mathrm{Precision}+\mathrm{Recall}} .
\end{equation}
When multiple successful reference trajectories are available for a task, we compute Tool F1 against each reference and use the best score. This avoids penalizing a model when it follows a valid alternative tool path.

\paragraph{Additional metrics.}
Besides Tool F1 and Task Success Rate, we compute several diagnostic metrics. \textbf{Next-tool Accuracy} measures whether the predicted next tool matches the reference next tool under a given context. \textbf{Average Steps} reports the average number of generated tool invocations. \textbf{Progress Rate} measures the fraction of examples where the model invokes at least one tool. \textbf{Argument Quality} evaluates whether generated tool arguments are compatible with the selected tool schema and historical observations. These metrics are used for diagnostics and are reported in Appendix~\ref{app:additional_main}.

\paragraph{Aggregation.}
For each method, we aggregate metrics over all valid held-out episodes. Tool F1, Next-tool Accuracy, Argument Quality, Progress Rate, and Task Success Rate are reported as percentages unless otherwise specified. For tables that average across settings, we first compute each metric within one dataset-backbone setting and then take the arithmetic mean over the six settings.

\begin{promptbox}{GPT-4o Judge System Prompt}
You are an expert evaluator of tool-use agents. Your task is to judge whether an agent's tool-use trajectory satisfies the user request.

Evaluation principles:
\begin{enumerate}
    \item Evaluate task completion, not exact imitation of the reference trajectory.
    \item The reference trajectory is one possible solution path. Alternative valid tools or orders should be accepted if they satisfy the task.
    \item Focus on whether the selected tools, arguments, observations, and final response support the user request.
    \item Minor tool-name differences, version suffixes, case differences, underscores, and hyphens should be ignored when the tool functions are equivalent.
    \item Do not penalize external simulator artifacts when the tool selection and arguments are reasonable.
    \item Penalize invalid arguments, missing required steps, unsupported conclusions, and premature stopping.
    \item For long-horizon tasks, partial completion may receive credit if the agent covers the main stages and makes meaningful progress toward the task.
\end{enumerate}
\end{promptbox}

\section{GPT-4o Judge Protocol}
\label{app:gpt4o_judge}

\begin{promptbox}{GPT-4o Judge User Prompt Template}
\textbf{User request}

\texttt{\{user\_request\}}

\medskip
\textbf{Reference tool trajectory}

The following trajectory is provided as a reference solution, not as the only valid solution:

\texttt{\{reference\_trajectory\}}

\medskip
\textbf{Agent tool trajectory}

Each step contains the tool name, arguments, and observation:

\texttt{\{agent\_trajectory\}}

\medskip
\textbf{Agent final response}

\texttt{\{final\_response\}}

\medskip
Please judge whether the agent completed the task. Output only a JSON object in the following format:

\begin{verbatim}
{
  "task_completed": "yes/partial/no",
  "score": <integer from 0 to 10>,
  "tool_selection_reasonable": <true/false>,
  "tool_order_reasonable": <true/false>,
  "uses_observations_correctly": <true/false>,
  "reason": "<brief reason in one or two sentences>"
}
\end{verbatim}

Scoring guideline:
\begin{itemize}
    \item 9--10: the trajectory completes the task and covers all key requirements.
    \item 7--8: the trajectory is mostly correct, with only minor omissions.
    \item 4--6: the trajectory makes meaningful progress and covers the main stages, but is incomplete.
    \item 1--3: the trajectory contains a few relevant invocations but misses core requirements.
    \item 0: no effective tool use is made.
\end{itemize}

Completion label:
\begin{itemize}
    \item \texttt{yes}: score \(\geq 7\).
    \item \texttt{partial}: score \(\geq 4\) and score \(<7\).
    \item \texttt{no}: score \(<4\).
\end{itemize}
\end{promptbox}

We use GPT-4o to evaluate Task Success Rate (TSR), because long-horizon tool-use tasks often admit multiple valid tool trajectories. Exact matching against a single reference path may underestimate success when the agent uses functionally equivalent tools or follows a different valid order. The judge therefore evaluates whether the generated trajectory satisfies the user request, instead of requiring exact reference imitation.

\paragraph{Judge model and inputs.}
We use GPT-4o with temperature \(0.1\) and a maximum generation length of 512 tokens. For each episode, the judge receives the user request, a reference tool trajectory, the agent's generated tool trajectory, truncated tool observations, and the final response. Method names are hidden from the judge. The same judge model and prompt are used for all methods. To control prompt length, we truncate each tool observation to 500 characters, the user request to 1000 characters, and the final response to 500 characters.

\paragraph{Evaluation principle.}
The judge evaluates task completion rather than surface-level similarity to the reference trajectory. The reference trajectory is treated as one possible solution path, not the only correct path. Functionally equivalent tools, minor naming differences, and different but reasonable invocation orders are accepted when they satisfy the same task requirement. Invalid tool arguments, missing required information, unsupported final answers, or premature stopping are treated as failures. External simulator artifacts, such as empty returns or API-like failures, are not penalized when the selected tool and arguments are otherwise reasonable.

\paragraph{Completion labels.}
The judge outputs one of three labels: \texttt{yes}, \texttt{partial}, or \texttt{no}. A trajectory is labeled \texttt{yes} if it completes the user request or covers all key requirements. It is labeled \texttt{partial} if it follows a reasonable tool-use strategy and completes the main stages of the task, but misses minor details or has incomplete final synthesis. It is labeled \texttt{no} if the tool trajectory is mostly irrelevant, misses core requirements, fails to use obtained observations, or stops before meaningful progress is made. In the main paper, TSR counts both \texttt{yes} and \texttt{partial} as successful:
\begin{equation}
\mathrm{TSR}
=
\frac{N_{\mathrm{yes}}+N_{\mathrm{partial}}}{N_{\mathrm{valid}}},
\end{equation}
where \(N_{\mathrm{valid}}\) is the number of episodes with a valid JSON judgment. We also record strict success rate using only \texttt{yes} for diagnostic analysis.

\paragraph{Parsing and aggregation.}
For each episode, we parse the judge response as JSON. If the response cannot be parsed, the episode is marked as an evaluation error and excluded from the valid denominator. For each method, we aggregate judge outputs over all valid held-out episodes and report TSR in the main table.

\section{Baseline Details}
\label{app:baseline_details}

This section describes how we adapt baseline methods to our long-horizon tool-use setting. 
All baselines are evaluated with the same held-out split, tool registry, tool simulator, maximum step budget, and evaluation protocol as CITA. 
For fair comparison, all trainable baselines start from the same SFT-initialized backbone used by CITA. 
Prompting and search-based baselines use the same tool invocation format and tool list. 
RL-based baselines use the same rollout budget and GRPO implementation unless the original method specifies a different optimization rule. 
Predictive and planning-based baselines use the same candidate generation budget as CITA whenever candidate-level comparison is required.

\subsection{Prompting and Search-Based Tool Use}

\paragraph{ReAct.}
ReAct interleaves natural-language reasoning with tool actions. 
We implement ReAct by adding a method-specific instruction that asks the model to alternate between a short reasoning step and a structured tool invocation. 
The tool invocation is still required to follow the same format used by all methods:
\[
\texttt{<tool\_call>\{"name": "...", "arguments": \{...\}\}</tool\_call>}.
\]
After each tool observation, the model updates its reasoning state and either invokes another tool or produces a final answer. 
No CIM score or comparative reward is used.

\paragraph{ToolChain*.}
ToolChain* formulates tool use as action-space navigation and applies A* search to find a low-cost tool-use path. 
We adapt this idea by building a search tree over candidate tool invocations. 
Each node represents a partial tool-use trajectory, and each outgoing edge corresponds to a possible next tool invocation generated by the policy. 
We score each partial path with a cost that combines model generation cost, tool-name validity, argument validity, and task-progress heuristic. 
The search keeps a fixed frontier size and expands candidates until the model terminates or reaches the maximum step limit. 
The same tool simulator is used to execute expanded tool invocations. 
This baseline tests whether explicit search over tool-use paths can replace CIM-based value estimation.

\paragraph{ToolLLM.}
ToolLLM provides a general framework for tool-use data construction, model training, and tool invocation evaluation. 
Since the original ToolLLM uses ToolBench and its own tool environment, we implement a ToolLLM-style baseline in our benchmark environment. 
The model is initialized from the same SFT checkpoint and uses the same tool descriptions. 
At inference time, we use a decision-tree search strategy over generated tool invocations, following the ToolLLM setting where the agent explores tool-use trajectories and selects a valid execution path. 
The search budget, maximum step limit, tool formatting rules, and simulator are matched to the other methods.

\subsection{Reinforcement Learning and Step-Level Reward Baselines}

\paragraph{ToolRL.}
ToolRL studies reward design for tool learning under reinforcement learning. 
We adapt ToolRL by training the SFT policy with GRPO using tool-use rewards that do not include CIM. 
The reward contains task-level trajectory matching, tool validity, and argument validity terms. 
This baseline uses the same rollout generation budget and held-out split as CITA, but it does not compare comparative next-tool choices under the same context.

\paragraph{StepTool.}
StepTool introduces step-grained reinforcement learning for multi-step tool use. 
We implement this baseline by adding step-level rewards derived from reference tool trajectories. 
At each step, the model receives feedback based on whether the selected tool matches the corresponding reference step or supports progress toward a reference trajectory. 
The step-level rewards are aggregated with the task-level reward for GRPO training. 
Unlike CITA, StepTool does not train a separate comparative value model and does not score alternative candidate invocations before execution.

\paragraph{IPR.}
IPR performs iterative step-level process refinement by estimating step-level rewards and using them to improve agent behavior. 
We adapt IPR by generating alternative actions along reference prefixes and evaluating them with rollout-based or reference-based feedback. 
The resulting step-level comparisons are used to refine the policy iteratively. 
To keep the comparison fair, IPR uses the same SFT initialization, tool simulator, maximum step budget, and rollout budget as CITA. 
The key difference is that IPR estimates process feedback through iterative refinement, whereas CITA trains CIM to directly estimate the long-horizon value of possible next-tool invocations.

\paragraph{AWPO.}
AWPO integrates reasoning-related reward signals into policy optimization through adaptive advantage weighting. 
We adapt AWPO by combining the task-level reward with auxiliary reasoning and format rewards, then applying advantage weighting during GRPO training. 
The reasoning reward encourages coherent task decomposition and valid tool-use planning, while the format reward encourages valid structured tool invocations. 
No CIM score, Bayesian simulator pair, or LLM-based comparative pair is used in this baseline.

\paragraph{AgentPRM.}
AgentPRM trains a process reward model for agent decisions by modeling both promise and progress. 
We adapt AgentPRM by training a step-level reward model from the same logged trajectories used in our experiments. 
The reward model scores intermediate tool-use states and candidate actions according to their estimated progress toward task completion. 
During policy optimization, AgentPRM scores are used as step-level rewards or candidate-ranking signals. 
This baseline evaluates whether a general process reward model for agents can substitute for CIM. 
Unlike CIM, AgentPRM is not trained on the three-source paired comparative construction and does not explicitly match value gaps between alternative tool invocations under the same context.

\subsection{Predictive and Planning-Based Baselines}

\paragraph{DyMo.}
DyMo learns an internal environment model for stateful tool use by combining function calling with state prediction. 
We adapt DyMo by adding a state-prediction objective to the SFT policy. 
After each tool invocation, the model predicts a compact summary of the resulting state before observing or using the next tool. 
At inference time, candidate invocations are ranked according to their predicted state progress and tool validity. 
This baseline tests whether implicit state-dynamics prediction can provide enough lookahead for long-horizon tool selection.

\paragraph{WebDreamer.}
WebDreamer uses an LLM as a world model and value function to simulate outcomes of candidate actions before execution. 
We adapt this idea by prompting an LLM planner to imagine the likely observation and task progress after each candidate tool invocation. 
The imagined outcomes are then scored, and the highest-scoring candidate is executed in the real tool simulator. 
The number of candidates and the maximum step budget are matched to CITA. 
Unlike CITA, WebDreamer-style planning relies on natural-language outcome simulation at inference time rather than a trained CIM that directly estimates next-tool value.

\paragraph{WKM.}
WKM provides global task knowledge before execution and local dynamic knowledge during agent planning. 
We adapt WKM by first generating a task-level knowledge summary that describes likely subtasks, relevant tools, and expected intermediate states. 
During execution, the model updates a local state summary after each tool observation and conditions the next tool invocation on both the global and local knowledge. 
This baseline tests whether explicit planning knowledge can improve tool-use trajectories without learning a comparative value estimator.

\subsection{Fairness and Budget Control}

For all baselines, we control the following factors. 
First, all methods use the same tool registry and the same per-task tool list construction. 
Second, all generated tool invocations are executed by the same simulator and are subject to the same maximum step limit. 
Third, trainable methods use the same SFT initialization and held-out split. 
Fourth, methods that require candidate comparison use the same candidate budget as CITA whenever possible. 
Finally, all methods are evaluated with the same Tool F1 computation and the same GPT-4o judge protocol for Task Success Rate. 
These controls ensure that performance differences mainly reflect the baseline method rather than differences in backbone, tool access, evaluation, or decoding budget.

\section{Additional Results and Confidence Intervals}
\label{app:additional_main}

This section reports additional diagnostic metrics for CITA across benchmarks and backbone models. 
The main text reports Tool F1 and Task Success Rate (TSR). 
Here we also include tool-level precision, tool-level recall, Next-tool Accuracy, Progress Rate, Argument Quality, and average number of generated tool invocations. 
For the two main metrics, Tool F1 and TSR, we report \(95\%\) confidence intervals estimated by task-level bootstrap resampling.

\paragraph{Metrics.}
We report additional diagnostic metrics to characterize different aspects of tool-use behavior. 
\textbf{Precision} measures the fraction of generated tool invocations that match a reference trajectory, and reflects whether the agent avoids unnecessary or irrelevant tools. 
\textbf{Recall} measures the fraction of reference tool invocations covered by the generated trajectory, and reflects whether the agent includes the tools needed for task completion. 
\textbf{Tool F1} is the harmonic mean of precision and recall:
\begin{equation}
\mathrm{F1}
=
\frac{2\cdot \mathrm{Precision}\cdot \mathrm{Recall}}
{\mathrm{Precision}+\mathrm{Recall}} .
\end{equation}
\textbf{TSR} measures whether the completed trajectory satisfies the user request according to the GPT-4o judge protocol in Appendix~\ref{app:gpt4o_judge}. 
\textbf{Next-tool Accuracy} measures whether the predicted next tool matches the reference next tool under the current context. 
\textbf{Progress Rate} measures the fraction of examples where the agent invokes at least one tool, indicating whether the model actually enters the tool-use process. 
\textbf{Argument Quality} measures whether generated tool arguments are compatible with the selected tool schema and the observed task state. 
\textbf{Avg. Steps} reports the average number of tool invocations in generated trajectories, which helps diagnose whether a method tends to stop too early or generate overly long trajectories. 
Tool F1 and TSR are the primary metrics, while the remaining metrics provide diagnostic evidence for tool selection, tool-use engagement, argument validity, and trajectory length.

\begin{table*}[ht]
\centering
\scriptsize
\setlength{\tabcolsep}{3.2pt}
\renewcommand{\arraystretch}{1.08}
\caption{Additional results and confidence intervals for CITA across benchmarks and backbones. Tool F1 and TSR are reported with estimated \(95\%\) confidence intervals.}
\label{tab:additional_main_results}
\resizebox{\textwidth}{!}{
\begin{tabular}{llcccccccc}
\toprule
Benchmark & Backbone 
& Prec. & Rec. 
& Tool F1 
& TSR 
& NTAcc. 
& Prog. 
& ArgQ. 
& Avg. Steps \\
\midrule
TOOLATHLON & Qwen2.5-7B 
& 66.20 & 52.50 
& \(58.56 \pm 1.62\) 
& \(47.82 \pm 1.95\) 
& 44.12 & 100.00 & 72.12 & 8.93 \\
TOOLATHLON & Llama3.1-8B 
& 65.40 & 54.22 
& \(59.29 \pm 1.75\) 
& \(45.05 \pm 2.03\) 
& 49.58 & 100.00 & 67.15 & 8.26 \\
TOUCAN & Qwen2.5-7B 
& 63.80 & 56.04 
& \(59.67 \pm 1.88\) 
& \(54.23 \pm 2.22\) 
& 55.00 & 100.00 & 55.08 & 2.50 \\
TOUCAN & Llama3.1-8B 
& 64.10 & 50.42 
& \(56.44 \pm 1.76\) 
& \(47.08 \pm 2.05\) 
& 60.00 & 100.00 & 62.58 & 2.05 \\
TRAJECT-Bench & Qwen2.5-7B 
& 66.17 & 53.72 
& \(59.30 \pm 1.64\) 
& \(48.31 \pm 2.10\) 
& 60.00 & 100.00 & 23.60 & 4.80 \\
TRAJECT-Bench & Llama3.1-8B 
& 75.30 & 64.67 
& \(69.58 \pm 1.42\) 
& \(55.72 \pm 1.84\) 
& 62.50 & 100.00 & 32.64 & 5.00 \\
\midrule
Average & -- 
& 66.83 & 55.26 
& \(60.47 \pm 1.68\) 
& \(49.70 \pm 2.03\) 
& 55.20 & 100.00 & 52.20 & 5.26 \\
\bottomrule
\end{tabular}
}
\end{table*}

The additional metrics provide a more detailed view of CITA's behavior. 
The precision and recall values are balanced across most settings, indicating that CITA does not simply increase recall by generating many extra tool invocations, nor does it over-prune trajectories at the cost of missing key tools. 
The high Progress Rate shows that CITA reliably enters the tool-use process rather than stopping before taking action. 
Next-tool Accuracy is lower than Tool F1 in several settings, which is expected because multiple valid tool trajectories may solve the same task and a single reference next tool can be too strict. 
The average number of steps also follows the benchmark profiles: Toolathlon requires longer trajectories, TOUCAN is shorter, and TRAJECT-Bench lies between them. 
Overall, these diagnostics support the main results: CITA improves trajectory-level tool selection while maintaining effective end-to-end task completion.

\section{Confidence Calibration of CIM}
\label{app:confidence_calibration}

We further evaluate whether the confidence head of CIM produces a meaningful reliability estimate. 
In our method, the confidence score \(c_\phi(s,a)\) is used together with the value estimate \(V_\phi(s,a)\) during candidate scoring and reward aggregation. 
This experiment tests whether a higher predicted confidence indeed corresponds to a higher empirical prediction accuracy.

For each held-out context-action pair \((s,a)\), we record the predicted value \(V_\phi(s,a)\) and confidence score \(c_\phi(s,a)\). 
We convert the value prediction into a binary success decision by thresholding at 0.5. 
That is, the model predicts success when \(V_\phi(s,a) > 0.5\), and predicts failure otherwise. 
This binary prediction is then compared with the ground-truth binary label to determine whether the prediction is correct.

To evaluate calibration, we partition all examples into 10 bins according to the predicted confidence, from \([0,0.1)\) to \([0.9,1.0]\). 
For each bin, we compute the average confidence and the empirical accuracy. 
A perfectly calibrated model should satisfy that the average confidence matches the empirical accuracy in each bin. 
We summarize calibration quality using Expected Calibration Error (ECE), defined as

\begin{equation}
\mathrm{ECE}
=
\sum_{b=1}^{B}
\frac{n_b}{N}
\left|
\mathrm{acc}(b)-\mathrm{conf}(b)
\right|,
\end{equation}

where \(B\) is the number of bins, \(n_b\) is the number of examples in bin \(b\), \(N\) is the total number of examples, \(\mathrm{acc}(b)\) is the empirical accuracy of bin \(b\), and \(\mathrm{conf}(b)\) is the average predicted confidence in bin \(b\).

\begin{figure*}[ht]
  \centering
  \includegraphics[width=\textwidth]{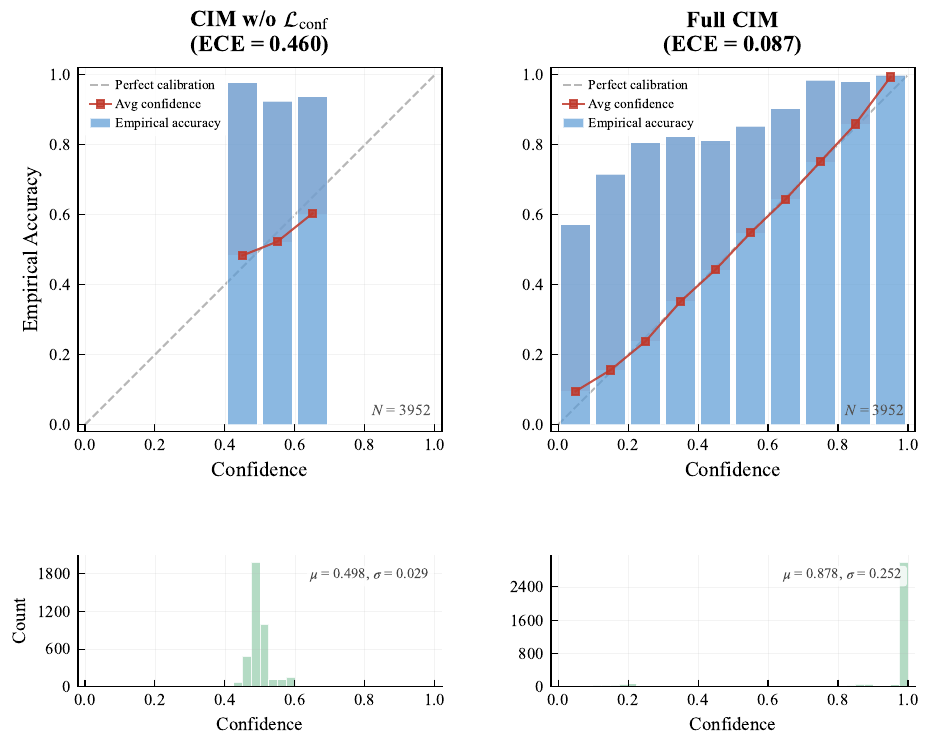}
\caption{
Confidence calibration of CIM. 
We compare a CIM variant trained without the confidence loss \(\mathcal{L}_{\mathrm{conf}}\) and the full CIM on held-out context-action pairs. 
In the upper row, blue bars show empirical accuracy in each confidence bin, the red line shows average predicted confidence, and the dashed diagonal indicates perfect calibration. 
In the lower row, histograms show the distribution of predicted confidence scores. 
The full CIM achieves much better calibration, reducing ECE from 0.460 to 0.087.
}
  \label{fig:confidence_calibration}
\end{figure*}

Figure~\ref{fig:confidence_calibration} compares two versions of CIM: a variant trained without the confidence loss \(\mathcal{L}_{\mathrm{conf}}\), and the full CIM. 
The upper row shows reliability diagrams, where the blue bars denote empirical accuracy in each confidence bin, the red line denotes average confidence, and the diagonal dashed line denotes perfect calibration. 
The lower row shows the distribution of predicted confidence scores.

The results show a clear difference between the two variants. 
Without \(\mathcal{L}_{\mathrm{conf}}\), the confidence predictions concentrate in a narrow middle range, with mean \(0.498\) and standard deviation \(0.029\). 
Its ECE is \(0.460\), and the empirical accuracy in those bins is much higher than the predicted confidence. 
This indicates that the model is poorly calibrated and systematically under-confident. 
In contrast, the full CIM produces a much wider confidence distribution, with mean \(0.878\) and standard deviation \(0.252\), and reduces ECE to \(0.087\). 
Its average confidence follows the empirical accuracy much more closely across bins, especially in the high-confidence region where most examples are located.

These results support the role of \(\mathcal{L}_{\mathrm{conf}}\) in CITA. 
The confidence head is not merely an auxiliary output. 
Instead, it learns a useful reliability signal that better reflects whether a value prediction is likely to be correct. 
This finding is consistent with the ablation results in the main text, where removing confidence-based scoring leads to lower Tool F1 and task success.

\section{Computational Cost}
\label{app:cost}

This section reports the computational cost of CITA during inference. 
All measurements are conducted on the same hardware with 2 H100 GPUs. 
We measure wall-clock inference time per task, excluding one-time model loading and dataset I/O, but including decoding, candidate generation, CIM scoring, tool execution, and context updates. 
All methods use the same held-out split, tool simulator, maximum step budget, and evaluation protocol.

\paragraph{Compared methods.}
We compare representative methods with different inference patterns. 
SFT uses direct greedy decoding after supervised fine-tuning. 
ToolRL and StepTool use RL-tuned policies with greedy decoding at inference time. 
WebDreamer represents planning-based inference with online future-state simulation. 
WKM augments the policy with global and local planning knowledge. 
CITA samples multiple candidate tool invocations at each step, scores them with the frozen CIM, and executes the highest-scoring invocation. 
In the main setting, CITA uses \(K=3\) candidates per decision step.

\paragraph{Cost metrics.}
Besides inference time per task, we also report relative inference time and time per successful task. 
The latter measures the expected wall-clock time needed to obtain one successful trajectory:
\begin{equation}
\mathrm{TimePerSuccess}
=
\frac{\mathrm{TimePerTask}}{\mathrm{TSR}/100}.
\end{equation}
This metric accounts for both efficiency and task success, and is therefore more informative than raw latency alone for long-horizon tool-use agents.

\begin{table}[ht]
\centering
\small
\setlength{\tabcolsep}{4.5pt}
\renewcommand{\arraystretch}{1.08}
\caption{Inference cost and task success on the same 2 H100 hardware. TSR is averaged over the six dataset-backbone settings. Time per successful task is computed from average inference time and TSR.}
\label{tab:computational_cost}
\begin{tabular}{lcccc}
\toprule
Method & Inference strategy & Time / task (s) \(\downarrow\) & TSR (\%) \(\uparrow\) & Time / success (s) \(\downarrow\) \\
\midrule
SFT & Greedy decoding & \(3.10 \pm 0.08\) & 23.84 & 13.00 \\
ToolRL & RL policy, greedy decoding & \(3.16 \pm 0.09\) & 26.60 & 11.88 \\
StepTool & Step-RL policy, greedy decoding & \(3.22 \pm 0.10\) & 26.59 & 12.11 \\
WKM & Knowledge-guided planning & \(4.65 \pm 0.14\) & 36.41 & 12.77 \\
WebDreamer & Online future-state simulation & \(11.92 \pm 0.36\) & 35.11 & 33.95 \\
CITA & Actor candidates + CIM scoring & \(4.84 \pm 0.16\) & \textbf{49.70} & \textbf{9.74} \\
\bottomrule
\end{tabular}
\end{table}

\paragraph{Analysis.}
Table~\ref{tab:computational_cost} shows that CITA introduces moderate inference overhead compared with greedy decoding. 
This overhead comes from sampling multiple candidate tool invocations and scoring them with CIM at each decision step. 
However, the additional cost is small compared with methods that perform online future-state simulation, such as WebDreamer. 
CITA is \(2.46\times\) faster than WebDreamer in average inference time, because CIM directly estimates next-tool value with a trained value model instead of repeatedly simulating future trajectories with an LLM planner.

The higher success rate also changes the cost comparison. 
Although CITA is slower than SFT, ToolRL, and StepTool in raw inference time, it achieves a much higher average TSR. 
As a result, CITA obtains the lowest time per successful task, reducing the expected cost of obtaining a successful trajectory to 9.74 seconds. 
This indicates that the additional CIM scoring cost is compensated by the improved decision quality.

\paragraph{Offline cost.}
CITA also introduces offline costs for paired data construction and CIM training. 
The Bayesian simulator pairs are inexpensive because they are generated on the tool transition graph without executing real APIs. 
The LLM-based comparative pairs are more costly, but they are constructed offline and reused throughout CIM training and policy optimization. 
At inference time, CITA does not call an LLM judge or perform LLM-based comparative rollout imagination. 
It only uses the trained CIM to score candidate invocations. 
This amortizes the cost of comparative supervision and keeps online inference substantially cheaper than planning methods that simulate future states at every decision step.

\paragraph{Scalability.}
The main online overhead of CITA scales linearly with the number of sampled candidates \(K\) and the number of decision steps. 
In practice, we use a small candidate budget, \(K=3\), which is sufficient to provide alternatives for CIM comparison while keeping latency moderate. 
The \(K=1\) ablation in Appendix~\ref{app:ablation} shows that removing candidate comparison hurts performance, confirming that a small amount of additional computation is useful for long-horizon tool selection.

\section{Case Study: CIM-Guided Tool Selection}
\label{app:case_study}

We provide two qualitative examples from Toolathlon with the Qwen2.5-7B backbone. 
Each example compares greedy SFT decoding with CITA at a critical decision point. 
For CITA, the actor proposes \(K=3\) candidate next-tool invocations, and CIM scores each candidate by its value estimate and confidence:
\begin{equation}
S_\phi(s,a)=V_\phi(s,a)c_\phi(s,a).
\end{equation}
The selected invocation is the candidate with the highest \(S_\phi(s,a)\).

\subsection{Example 1: Canvas Quiz Completion}

\paragraph{Task.}
\textit{``My personal information is stored in memory. Check which unfinished course quizzes I have on Canvas, and help me complete them.''}

\paragraph{Reference path.}
\texttt{health\_check}
\(\rightarrow\)
\texttt{list\_courses}
\(\rightarrow\)
\texttt{list\_quizzes}
\(\rightarrow\)
\texttt{start\_quiz\_attempt}
\(\rightarrow\)
\texttt{list\_quiz\_questions}
\(\rightarrow\)
\texttt{submit\_quiz\_answers}.

This task has a strict prerequisite: the agent must start a quiz attempt before it can list questions or submit answers. 
The SFT baseline correctly invokes \texttt{health\_check}, \texttt{list\_courses}, and \texttt{list\_quizzes}, but then jumps directly to \texttt{list\_quiz\_questions}. 
Because no active quiz attempt exists, the subsequent quiz operations fail.

\begin{table}[ht]
\centering
\small
\setlength{\tabcolsep}{5pt}
\renewcommand{\arraystretch}{1.05}
\caption{CIM scores at the critical fork point in the Canvas quiz example.}
\label{tab:case_canvas}
\begin{tabular}{lccc}
\toprule
Candidate next-tool invocation & \(V_\phi(s,a)\) & \(c_\phi(s,a)\) & \(S_\phi(s,a)\) \\
\midrule
\textbf{\texttt{start\_quiz\_attempt}} & \textbf{0.74} & \textbf{0.82} & \textbf{0.61} \\
\texttt{list\_quiz\_questions} & 0.28 & 0.71 & 0.20 \\
\texttt{get\_quiz} & 0.33 & 0.65 & 0.21 \\
\bottomrule
\end{tabular}
\end{table}

CIM assigns the highest score to \texttt{start\_quiz\_attempt}, the prerequisite-enabling tool. 
Although \texttt{list\_quiz\_questions} is locally plausible after identifying an unfinished quiz, its low value reflects that it leads to an invalid state without an active attempt. 
This example shows how CIM-guided tool selection can avoid a locally plausible but downstream-invalid invocation.

\subsection{Example 2: Trip Planning with Multiple Services}

\paragraph{Task.}
\textit{``I am currently at the UPenn Bookstore. I would like to take a walking tour of campus attractions. A few days ago, I saved the list of sights in a workspace file. Please read it, plan a route, and write the result.''}

\paragraph{Reference path.}
\texttt{read\_file}
\(\rightarrow\)
\texttt{geocode}
\(\rightarrow\)
\texttt{distance\_matrix}
\(\rightarrow\)
\texttt{directions}
\(\rightarrow\)
\texttt{write\_file}
\(\rightarrow\)
\texttt{claim\_done}.

This task requires cross-service coordination. 
The agent must first read the saved sights, then convert locations into coordinates before computing distances and directions. 
The SFT baseline reads the file but skips \texttt{geocode}, then invokes map tools with raw place names. 
This makes the following location-based invocations unreliable and prevents the agent from writing a complete route plan.

\begin{table}[ht]
\centering
\small
\setlength{\tabcolsep}{5pt}
\renewcommand{\arraystretch}{1.05}
\caption{CIM scores at the key map-planning fork point in the trip planning example.}
\label{tab:case_trip}
\begin{tabular}{lccc}
\toprule
Candidate next-tool invocation & \(V_\phi(s,a)\) & \(c_\phi(s,a)\) & \(S_\phi(s,a)\) \\
\midrule
\textbf{\texttt{geocode}} & \textbf{0.71} & \textbf{0.79} & \textbf{0.56} \\
\texttt{search\_places} & 0.38 & 0.62 & 0.24 \\
\texttt{directions} & 0.25 & 0.55 & 0.14 \\
\bottomrule
\end{tabular}
\end{table}

CIM selects \texttt{geocode} before route planning, assigning it a higher value than \texttt{search\_places} or \texttt{directions}. 
This reflects the long-horizon dependency that coordinates are needed before distance and direction queries become reliable. 
After geocoding, CITA proceeds with \texttt{distance\_matrix}, \texttt{directions}, and \texttt{write\_file}, producing a coherent tool-use trajectory.

Both examples illustrate the same pattern. 
Greedy decoding selects tools that are locally plausible but violate downstream dependencies. 
CIM-guided tool selection instead compares candidate next-tool invocations by their estimated long-horizon value under the current context. 
Prerequisite-enabling tools receive higher value and confidence, while candidates that lead to dead-end states receive lower combined scores. 
This helps CITA connect step-level tool selection with final task success.

\section{Broader Impact and Responsible Use}
\label{app:broader_impact}

CITA aims to improve the reliability of long-horizon tool-use agents by helping them compare candidate next-tool invocations before execution. 
More reliable tool selection may reduce unnecessary invocations, improve task completion, and make agent behavior easier to inspect through value and confidence scores.

At the same time, stronger tool-use agents should be deployed with appropriate safeguards. 
External tools may access files, APIs, databases, or other stateful environments, where incorrect invocations can have undesirable effects. 
CITA improves decision quality, but it does not replace environment-level safety controls. 
For real deployment, tool permissions, sandboxing, rate limits, execution logs, and human confirmation for irreversible or high-impact actions remain important.

Our experiments are conducted in benchmarked tool environments rather than real-world high-stakes deployments. 
For broader use, CIM-guided tool selection should be combined with policies that restrict sensitive tools, validate arguments before execution, and monitor agent behavior over time.



\end{document}